%% file: Main.tex
\documentclass[lettersize,journal]{cls/IEEEtran}

\IEEEoverridecommandlockouts		

\usepackage{graphicx}
\usepackage{amsmath} 
\usepackage{amssymb} 
\usepackage{cite}
\usepackage{amsfonts}
\usepackage{amsthm}
\usepackage{comment}
\usepackage[ruled,vlined,linesnumbered,noresetcount]{algorithm2e}

\usepackage{bm}
\usepackage{breqn}

\usepackage{footnote}
\makesavenoteenv{tabular}

\usepackage{caption}
\usepackage[numbers]{natbib}
\usepackage[table,usenames,dvipsnames]{xcolor}
\usepackage{enumitem}
\usepackage{graphicx,tabularx,adjustbox}
\usepackage{multicol}
\usepackage[font={small}]{caption}   
\usepackage{subcaption}
\usepackage{booktabs,makecell}

\usepackage[subtle,tracking=normal]{savetrees} 

\makeatletter
\define@key{Gin}{mycrops}[]{\setkeys{Gin}{trim=10cm 2.5cm 5cm 0cm,clip}}
\makeatother

\newcommand{\add}[1]{\textcolor{black}{#1}}
\newcommand{\NEW}[1]{\textcolor{black}{#1}}

\usepackage{soul}

\usepackage{hyperref}
\usepackage[percent]{overpic}
\usepackage{xcolor}

\usepackage{multicol}
\usepackage{subcaption}

\begin{document}

\title{\LARGE \bf
 Mastering Agile Jumping Skills from Simple Practices \\ with Iterative Learning Control 
}

\author{Chuong Nguyen$^{*}$, Lingfan Bao$^{*}$, Quan Nguyen%
\thanks{
The authors are with the Dynamic Robotics and Control Laboratory, Department of Aerospace and Mechanical Engineering, University of Southern California (USC).
        {\tt\footnotesize vanchuong.nguyen@usc.edu, lingfanb@usc.edu,  quann@usc.edu}}%
\thanks{$^{*}$ Equally contributed to this work.}
}

\maketitle
\thispagestyle{empty}
\pagestyle{empty}

\input{Sections/abstract}

\input{Sections/content}

\bibliographystyle{IEEEtran}
\bibliography{chuong_ref} 

\end{document}

%% file: Sections/abstract.tex
\begin{abstract} 
\add{Achieving precise target jumping with legged robots poses a significant challenge due to the long flight phase and the uncertainties inherent in contact dynamics and hardware. Forcefully attempting these agile motions on hardware could result in severe failures and potential damage.}
Motivated by these challenging problems, we propose an Iterative Learning Control (ILC) approach that aims to learn and refine jumping skills from easy to difficult, instead of directly learning these challenging tasks. We verify that learning from simplicity can enhance safety and target jumping accuracy over trials. Compared to other ILC approaches for legged locomotion, our method can tackle the problem of a long flight phase where control input is not available. In addition, our approach allows the robot to apply what it learns from a simple jumping task to accomplish more challenging tasks within a few trials directly in hardware, instead of learning from scratch.
We validate the method via extensive experiments in the A1 model and hardware for various jumping tasks.  Starting from a small jump (e.g., a forward leap of $40cm$), our learning approach empowers the robot to accomplish a variety of challenging targets, including jumping onto a $20cm$ high box, jumping to a greater distance of up to $60cm$, as well as performing jumps while carrying an unknown payload of $2kg$.
Our framework can allow the robot to reach the desired position and orientation targets with approximate errors of $1 cm$
and $1^0$ within a few trials. 
\end{abstract}

%% file: Sections/content.tex
\section{Introduction}\label{sec:Introduction}


\NEW{Jumping is the unique capability of legged robots to navigate discrete terrain or cluttered spaces, with speed and efficiency \cite{AtlasBackflip2023,yaran_MICP_jump_2020,GabrielICRA2021,katz2019mini, fullbody_MPC,QuannICRA19,ZhitaoIROS22}.  
\textit{Target} jumping is crucial and has always been the primary goal when designing jumping controllers.
As we can observe in some motions such as leaping over gaps \cite{AtlasBackflip2023,yaran_MICP_jump_2020} or jumping onto high elevations \cite{RL_jump_bipedal,QuannICRA19}, even a small error in the landing position and pose could make the robot miss the landing surface. However, ensuring the accurate target jumping (i.e. target location and pose) is challenging \cite{yaran_MICP_jump_2020, RL_jump_bipedal}. The primary reason is that jumping typically involves long flight maneuvers. 
During this long aerial period, the system becomes under-actuated and the control has little impact on the robot's trajectory.  To successfully land on a given target, the robot needs to effectively coordinate the whole body and joints throughout the contact phase, as well as generate highly accurate translational and angular momentum at take-off \cite{yaranding_2018}.}



\begin{figure}[!t]
    \centering
\includegraphics[width=1\linewidth,trim={1.5cm 2cm 1cm 0cm},clip]{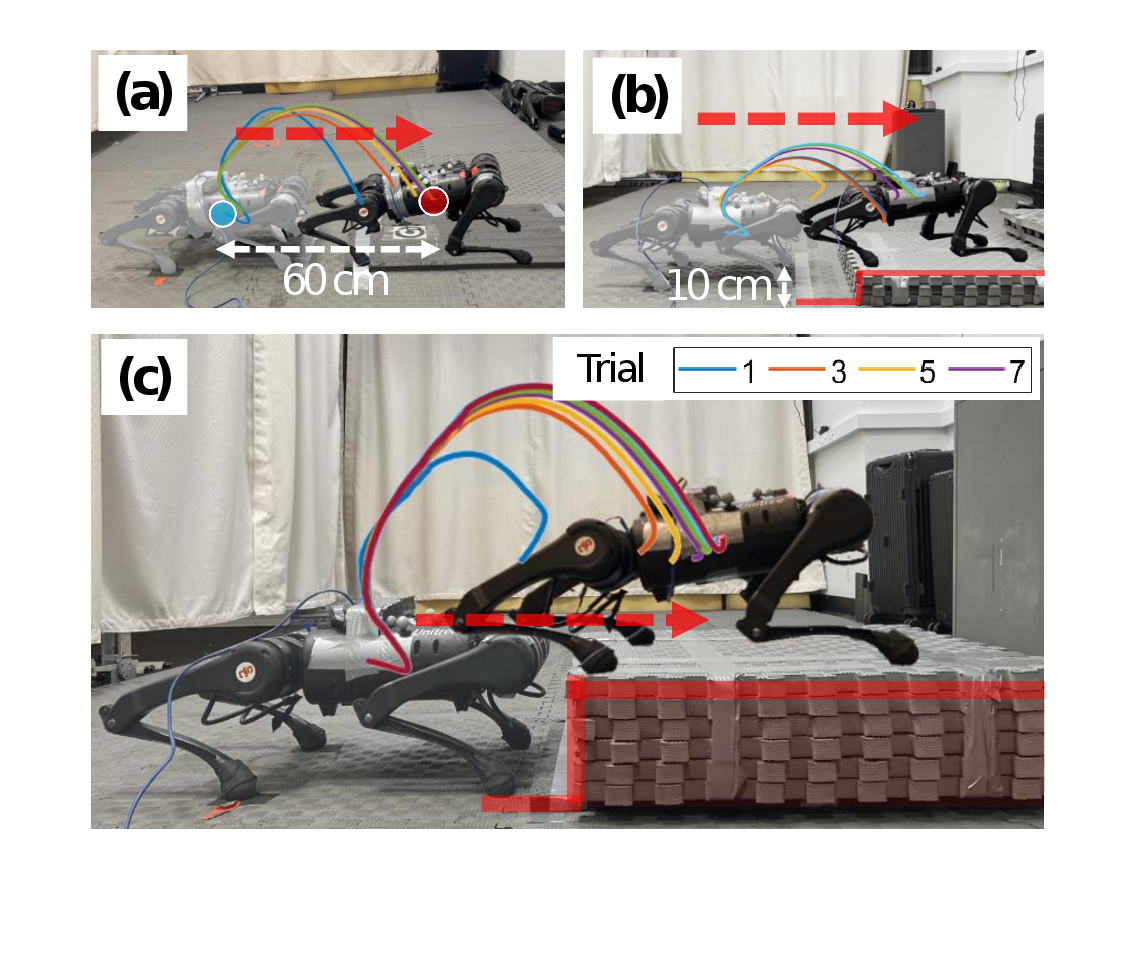}	
 \caption{\textbf{Learn and practice from easy to more challenging tasks.} Our approach enables the A1 robot to apply what it learned from a simple jump of $40~cm$ in order to accomplish more challenging tasks: (a) Jump farther to $60~cm$ within 9 trials, (b) jump on box at $(x,z)=(60, 10)~cm$ within 8 trials, (c) Jump on box at $(x,z)=(50, 20)~cm$ within 7 trials. Supplemental video: \textit{\url{https://youtu.be/zbEB5bMBgY0}}}
\label{fig:ILC_Introduction}
\end{figure}


\NEW{Another challenge of jumping practice is the \textit{Safety}. For agile and powerful jumps (e.g. box jumping or gap leaping), successfully landing on the target position might not be guaranteed at every first trial \cite{AtlasParkourFail}. Additionally, 
a large landing pose error combined with the significant body linear or angular velocity at the impact could make the robot keep rotating unexpectedly, which could challenge even recent advanced landing controllers to handle the hard impact and recover the robot safely \cite{quad_landing_nn, QuannICRA19,reactive_landing}.
In these scenarios, the landing normally fails badly, and the data collected might not even be helpful. Also, it might be inefficient to conduct many trials and adjust controller parameters until success, while incurring the cost of hardware damage. Lastly, it may be necessary to manually retune controller parameters for various tasks and goals, as indicated in \cite{YanranDingTRO}.}

Motivated by these challenges, we introduce a method based on Iterative Learning Control (ILC) that enhances target accuracy and safety in dynamic jumping. Instead of attempting more challenging tasks directly on legged robots, we propose to practice and master jumping skills on a simpler task before transferring the acquired knowledge to more challenging jumping tasks. One notable advantage of starting with short jumps is the enhanced safety and effectiveness of the learning process. Directly attempting to learn highly dynamic jumps may damage the robot and usually requires more time or trials to yield reasonable results or meaningful data (most attempts fail significantly, and the data collected are often not helpful). Our designed strategy is inspired by the training principle observed in athletes. Consider an athlete aiming for precise, long or high jumps: they initially practice simpler jumps and progressively refine their technique to achieve accurate targets and safe landings.

\subsection{Related Works}
This section on related work is divided into three parts. In the first two parts, we summarize recent model-based and learning-based approaches aimed at executing more challenging tasks directly. In the third part,  we discuss related works that have utilized iterative learning for trajectory tracking and legged locomotion.


\subsubsection{Model-based Approaches}
Recent advanced model-based approaches, such as Model Predictive Control (MPC), have been shown as effective in realizing various acrobatic motions in legged robots \cite{YanranDingTRO,GabrielICRA2021,continuous_jump, Quad-SDK,park2017high}. 
\add{However, since MPC approaches often depend on model simplification to ensure real-time execution feasibility, this can compromise both target accuracy and the success rate of transferring highly agile motions to hardware.} 
On the other hand, Trajectory optimization (TO) can improve target accuracy by accounting for the full-body dynamics model, making it another suitable method for planning acrobatic motions with extended flight times \cite{ matthew_mit2021_1,falling_cat,QuannICRA19}.
Nevertheless, achieving an accurate full-body model remains challenging, particularly for agile jumping maneuvers that introduce significant uncertainties and unmodeled dynamics. For example, the interaction between foot and ground introduces various uncertainties, such as varied friction or unknown ground stiffness and damping. Additionally, parameter variations, such as inaccurate kinematic parameters, also affect model accuracy. For instance, the authors in \cite{Zachary_RAL_online_calibration_2022} show that the actual leg length of the robot with deformable feet is likely difficult to measure due to dynamic deformation effects and rolling contacts. \NEW{These uncertainties and unmodelled dynamics, unfortunately, are normally simplified or ignored in trajectory optimization framework and real-time execution \cite{matthew_mit2021_1,falling_cat,QuannICRA19}. Therefore, it could result in errors in hardware transfer. 
In summary, despite impressive results, model-based approaches such as MPC or trajectory optimization may struggle to achieve accurate target jumping and may require numerous trials and parameter adjustments to succeed. Furthermore, when addressing other targets, they typically require new jumping references \cite{ matthew_mit2021_1,YanranDingTRO,continuous_jump}.
In this paper, we propose a method that allows the robot to practice jumping until it reaches a given target at high accuracy. In addition, our method enables the utilization of a single reference for various jumping tasks.}


\subsubsection{Learning-based approaches} Deep reinforcement learning has recently emerged as an attractive solution to realize agile motions on legged robots as well \cite{cheng2023parkour, Yuni_imitate_TO_quadrupedrobot,kang_modelbased+RL, RL_jump_bipedal,yang2023cajun,zhuang2023robot}.
\add{
An effective strategy involves integrating trajectory optimization, or more broadly, model-based optimal control techniques, into learning pipelines to refine demonstrations derived from trajectory optimization via RL
\cite{Yuni_imitate_TO_quadrupedrobot}, \cite{kang_modelbased+RL}. This approach has been validated to enable agile backflips \cite{Yuni_imitate_TO_quadrupedrobot} and enhance robustness against environmental uncertainties \cite{kang_modelbased+RL}. In our framework, we leverage control inputs from optimization as an initial point to initialize the learning process. However, we employ iterative learning control (ILC) to further refine jumping skills and enhance target jumping accuracy. Moreover, our ILC framework is designed to facilitate achieving multiple target jumps. A recent DRL solution toward this objective involves training with various targeted locations to exploit the diversity of learned maneuvers \cite{RL_jump_bipedal}}. 
\NEW{Nevertheless, DRL normally relies on extensive data collection and high-configuration computers for training. Moreover, due to the inherent complexity and time-consuming hardware experiments of jumping tasks, these DRL frameworks mainly rely on simulation to collect extensive
training data, which normally requires bridging the sim-to-real gap \cite{jump_mentor,RL_jump_bipedal,jumping_inpixel}. Different from DRL, our approach for jumping tasks requires less data and can efficiently solve for the optimal control inputs within a second in a standard computer. Additionally, our method enables robots to learn directly in hardware instead of heavily relying on simulation for hardware transfer as DRL does.
DRL also normally relies on random exploration of the control actions with extensive trials and errors until it converges to an optimal policy. Our proposed method, on the other hand, enforces control actions into an optimization framework instead of random exploration. Thus, it only requires a small number of trials to accomplish desired tasks, which is mostly unattainable through DRL.} 

\subsubsection{Iterative Learning Control}
To deal with uncertainties and unknown dynamics in repetitive tasks, Iterative Learning Control (ILC) offers an effective solution, as it allows the learning from failures and gradually improves the tracking accuracy performance over time ~\cite{HSAhn_ILC_survey, ILC_Chen2004, DDMPCILC,schoellig2012optimization}. This advantage gives ILC a wide range of applications for precise trajectory tracking such as manipulation control \cite{ILC_arm_ppmixed,DDMPCILC}, and quadcopter maneuver \cite{ DeeplearningILC,ilc_quadcopter,schoellig2012optimization}. 
Inversion methods invert plant dynamics for the learning function, as discussed in prior works \cite{bristow2006survey,invertILC_1999}. While exhibiting a rapid convergence rate, this approach can be sensitive
to model errors due to its reliance on modeling \cite{bristow2006survey}.
Without the need for an accurate model, a model-free PD-type ILC is widely utilized to track trajectory references, relying on tuning the PD learning gains \cite{bristow2006survey,ILC_Chen2004,Lee1998}. However, dynamic constraints are often ignored in this model-free approach. ILC can be integrated with MPC as a unified ILC-MPC framework, which can both tackle constraints and improve the tracking
performance \cite{DeweiLi2016,DDMPCILC,fuzzyILCMPC_2013}.
In the realm of ILC, most previous works focus on trajectory tracking and assume that control is always available during trajectory execution. However, jumping on legged robots poses a different unique problem that consists of a long flight time, and the controller is typically not applied during this period \cite{YanranDingTRO,GabrielICRA2021}.Thus, it is typically challenging to enable the robot to jump and land accurately at a given location.
In this paper, we design an ILC method that tackles long-flight maneuvers to enable accurate target jumping at the end.

Compared to the abundance of ILC work in robotics, the literature on the iterative learning of legged robots is comparatively sparse. Our most related works focus on walking gait stabilization for humanoids \cite{Kai_ILC_biped_ICRA15,Kai_ILC_TRO} and 2D pronking in quadrupeds \cite{pronking_ILC, ILC_continuous_pronking}. These works, however, consider locomotion with either no flight time or a short flight time.
In particular, the authors in \cite{Kai_ILC_biped_ICRA15,Kai_ILC_TRO} propose a framework using ILC to modify trajectory references, which then is to be tracked by a whole body controller to stabilize walking gaits in humanoid robots. In addition, the linear inverted pendulum (LIP) model is adopted to represent the slow dynamics of the walking motion. The approach in \cite{pronking_ILC} utilizes a full-body dynamic model for trajectory optimization framework to obtain reference joint profile. Once the joint reference is computed, a PD-type ILC is designed for joint trajectory tracking in simulation. The work demonstrates impressive joint tracking performance; however, the joint reference obtained from optimization assumes an accurate model of robot and contact dynamics, which may be challenging to guarantee in hardware. Additionally, since this method focuses solely on low-level joint tracking, achieving accurate jumping to a desired target is challenging, as this task typically requires high-level body tracking. Our proposed approach, however, focuses on the final jumping result rather than just the joint angle trajectory, as the latter does not effectively represent the jumping performance. The authors in \cite{ILC_continuous_pronking} alleviate the need to utilize the full-body dynamic model as \cite{pronking_ILC}, and instead use SRB to avoid heavy computation. They focus on trajectory tracking and adopt a functional ILC from \cite{ILC_functional} to enable effective pronking motions with a limited aerial maneuver ($\sim0.1s$).
Our work, on the other hand, aims to address the challenge of jumping motions with a long flight time and to tackle the problem of target jumping.
Additionally, unlike previous work on legged locomotion, we will allow the robot to practice with a simple jump and then use this experience to perform more challenging tasks quickly and safely.
Last but not least, our approach takes into account the true system limits, such as motor dynamic constraints, to enable dynamic jumping maneuvers in hardware.

\subsection{Contributions}

Our work aims to address some challenging problems for jumping on robot hardware as follows:
\begin{itemize}
    \item How to enable the robot to perform dynamic jumping maneuvers with accurate target and safety requirements?
    \item How to effectively leverage the learning progress obtained from a simple jump to accomplish more challenging tasks within several trials, instead of inefficiently learning from scratch?
\end{itemize}
To this end, we formulate jumping as a repetitive task, then propose a framework based on iterative learning control (ILC) as a potential solution to tackle these problems.
The main contributions of our work are outlined as follows: 
\begin{itemize}
    \item Our framework allows the robot to leverage its learning progress and skills obtained from a simple jumping task to accomplish other challenging tasks within a few trials. In addition, we demonstrate the feasibility of using a single reference for multiple goals.

    \item Our learning approach enhances both the target jumping accuracy and safety for jumping maneuvers with long aerial phases. It can allow the robot to reach the desired position and orientation targets with approximate errors of $1cm$ and $1^0$ within a few trials.


    \item We propose to integrate a model of motor dynamic constraints that represents a relationship between velocity and torque in DC motors. This integration aims to realize the hardware capabilities and enable successful learning directly in hardware.

\end{itemize}
The remainder of this manuscript is organized as follows. In
Section \ref{sec:ILC_Approach}, we first present the overview of the framework and propose to consider the motor dynamic constraints to represent the robot's system limits for jumping tasks. Section \ref{sec:ILC_Approach}.D presents the ILC framework to achieve accurate target jumping on a simple task, and to accomplish more challenging tasks based on what the robot learned from the simple practice. Section \ref{sec:ILC_Results} shows the comparisons with other ILC and MPC approaches, and verifies the proposed framework via extensive experiments in both simulation and hardware. The concluding remarks are
provided in Section \ref{sec:ILC_Conclusion}.

\begin{figure*}{
      \centering
      \includegraphics[trim = 0mm 0mm 0mm 2mm, clip, width=\textwidth]{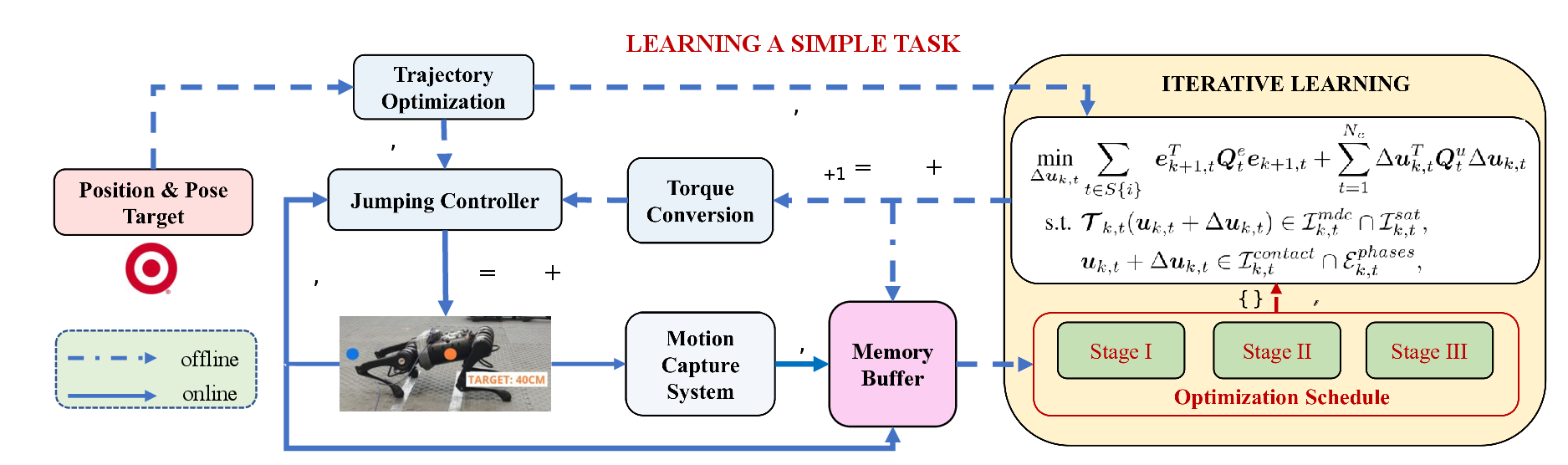} \\
      \caption{\textbf{Practicing a simple task.} This framework describes the learning process of a simple jumping maneuver\label{fig:overview_framework}
      \vspace{-0.5em}}}
\end{figure*}

\begin{figure*}{
      \centering
      \includegraphics[trim = 0mm 0mm 0mm 10mm, clip, width=\textwidth]{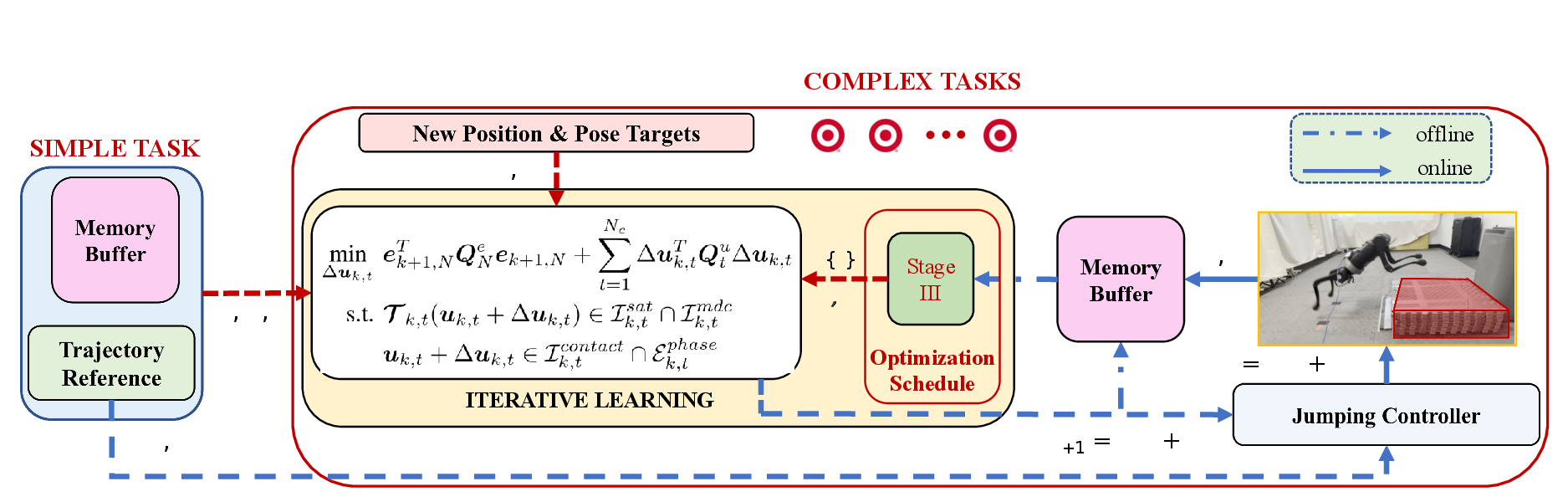} \\
      \caption{\textbf{Learning to complete further challenging tasks.}  Our proposed framework enables the learning from a simple task to more challenging goals within several trials, instead of learning from scratch. Joint reference profile of simple task $\bm{q}_d, \dot{\bm{q}}_d$ can be utilized for challenging tasks instead of re-running the trajectory optimization to get a new jumping reference}
      \label{fig:simple_to_complex_framework}}
\end{figure*}

\section{Proposed Approach}
\label{sec:ILC_Approach}

\subsection{Overview of the proposed framework}



Our proposed approach aims to facilitate a wide range of multiple-target jumping tasks, starting from simpler ones. This involves a learning procedure for a simple jump, illustrated in Fig. \ref{fig:overview_framework}. Subsequently, the process for mastering more challenging jumping tasks is detailed in Fig. \ref{fig:simple_to_complex_framework}. \add{In this context, the term 'simple task' or 'simple jump' refers to jumping a short distance, whereas 'challenging tasks' involve jumping farther or onto higher platforms.}


For learning a simple task with a given position and pose target, the trajectory optimization (TO) is formulated to solve for the nominal trajectory reference $\lbrace \bm{p}_d, \dot{\bm{p}_d}, \bm{q}_d, \dot{\bm{q}_d} \rbrace$ of the full-body dynamics, as detailed in section \ref{sec: ILC_TO}. The joint reference $\lbrace \bm{q}_d, \dot{\bm{q}_d} \rbrace$ is then used for the feedback joint PD controller, while $\lbrace \bm{p}_d, \dot{\bm{p}_d} \rbrace$ is utilized as body trajectory reference for the iterative learning framework. The memory buffer keeps the data record of body position and orientation at the current trial $k$, obtained from the motion capture system sampling at $1kHz$. Based on the data record, we compute the predicted trajectory error $\bm{e}_{k+1,t}$ of the whole trajectory $t\in [1, N]$ of the next trial. We propose to divide the learning process into three subsequent stages. Stage I leverages the learning for all-leg contact and rear-leg contact phase, followed by Stage II which aims to minimize trajectory errors for rear-leg contact and flight phases. The final stage III focuses on the target landing accuracy, minimizing the predicted error of the final position and pose $\bm{e}_{k+1,N}$ of the next trial $k+1$. For each stage, we formulate the iterative learning as an optimization problem to solve for the optimal offset of contact force $\Delta \bm{u}_{k}^*$ while satisfying all constraints related to hardware limits, friction cone, and contact schedule. The contact force $\bm{u}_{k}$ utilized during the trial $k$ is then added up with this optimized offset to apply to the next trial $\bm{u}_{k+1}=\Delta \bm{u}_{k}^* + \bm{u}_{k}$. The optimization formulation of the learning progress is presented in detail in Section \ref{sec:ILC_controller}2.  

\NEW{After successfully learning a simple task, we design a procedure to enable the learning to jump to more challenging goals, instead of learning from scratch.  
\add{The force control and trajectory of the robot's body and limbs (i.e., $\bm{u}_s, \bm{p}_s, \theta_s, \bm{q}_s$), learned from the simple task, will be used as an initialization for learning more challenging tasks.}
Given a new position and pose target $\bm{p}_{d}, \theta_{d}$, the ILC optimizes the control offset $\Delta \bm{u}_{k}^*$ and executes the first trial. By leveraging what the robot has learned from the simple task, we now only need to focus on Stage III, which aims to minimize the difference between predicted robot states at the end. We repeat the learning process in Stage III until the robot reaches the desired goal. The robot can efficiently accomplish more challenging tasks within several trials while enhancing safety during the learning process. 
}

\subsection{Motor Dynamic Constraint Modelling} \label{sec: ILC_MDC_modeling}

To accomplish aggressive jumping maneuvers, legged robots must reach their actuators and power limits rapidly. The whole jumping motion, which includes contact and flight phases, typically occurs within a short time frame of around $1\sim2$ seconds. It is crucial to consider these true system constraints when developing control strategies to minimize the gap in hardware transfer. Therefore, we propose to incorporate the motor dynamics constraints (MDC) into our ILC framework to realize these aggressive motions in hardware transfer.
The MDC represents the inherent torque-velocity relationship in conjunction with the supply limit (i.e., the on-board battery voltage).  To formulate this relation, we consider a simplified model for each DC motor that shows the voltage applied to each motor can be estimated as follows: 
\begin{align} \label{eq:voltage_cs}
V_i(\tau_i^m, \dot{q}_i^m)=I_i^m(\tau_i^m) r_i+\zeta_i(\dot{q}_i^m),
\end{align} 
where $r_i$ is the resistance of the coils windings, and $\dot{q}_i^m$ is the motor velocity. For this simplified model, 
\begin{itemize}
    \item We neglect the effect of inductance of stator windings because it is typically small (approximately $1 mH$ for an A1 robot motor \cite{unitreeA1}).
    \item The back electromotive force (EMF) of the windings generated by the rotation of the motor is estimated by $\zeta_i(\dot{q}_i^m) = K_v\dot{q}_i^m$. Here, $K_v$ is the electric motor velocity constant.
    \item The current $I_i^m(\tau_i^m)$ flowing in the windings relates to the motor torque via $I_i^m = \tau_i^m /K_{\tau} $. Here, $K_{\tau}$ is the torque constant. 
\end{itemize}

Considering the gear ratio $g_r$ which relates $\tau_i = \tau_i^m g_r$ and $\dot{q}_i = \dot{q}_i^m/ g_r$, we can rewrite the voltage equation (\ref{eq:voltage_cs}) as a linear combination of joint torque and joint velocity as
\begin{align} \label{eq:}
V_i(\tau_i, \dot{q}_i)=\rho \tau_i+ \sigma \dot{q}_i,
\end{align} 
where $\rho =r_i/(K_\tau g_r)$ and $\sigma =K_vg_r$, respectively. MDC establishes a key relation between joint torque and velocity in conjunction with the available supply voltage  $V_{bat}$, i.e.,
    \begin{align} \label{eq:MDC_voltage}
        | V_i(\tau_i, \dot{q}_i)|=\lvert \rho \tau_i+ \sigma \dot{q}_i \rvert \leq V_{bat}
    \end{align}
The MDC (\ref{eq:MDC_voltage}) states that the joint torques and joint velocities cannot both reach their limits at the same time. This mean that, for example, the DC motor can attain its maximum velocity only when running at no load, and the back EMF approaches the supply voltage. 
Approximately, $\dot{q}_i^{max}=V_{bat}/\sigma$, giving rise to the following constraints: 
\begin{align} \label{eq:max_vel}
    V_{bat} \geq  \sigma \dot{q}_i, ~ -V_{bat} \leq  \sigma \dot{q}_i
\end{align}






\subsection{Trajectory Optimization}
\label{sec: ILC_TO}
We utilize trajectory optimization to generate jumping references for a simple task.  It is worth noting that the proposed ILC method allows us to use
the reference trajectory of a simple task to leverage more challenging tasks. This omits the need to rerun optimization for different
targets.
The optimization framework adopts the full-body dynamics of the robot to leverage the whole-body motion for jumping. The optimization formulation is similar to our prior work \cite{QuannICRA19}.
This considers the generalized jumping tasks as having three distinct contact phases: all-leg contact phase, rear-leg contact phase, and the subsequent flight phase. These phases are denoted as $dc$, $sc$,$fl$, and take $N_{dc}$, $N_{sc}$, and $N_{fl}$ time steps, respectively. Then, the resulting discrete-time optimization can be formulated as follows:

\begin{align}
 \textrm{ }&\text{minimize} \textrm{ } J\bigl(\bm{q}_N\bigr) + \delta t \sum_{k=1}^{N-1} w\bigl(\bm{q}_k,\bm{\tau}_k\bigr) \nonumber\\
\mbox{s.t} \ & \lambda \bigl(\bm{s}_k,\bm{s}_{k+1}, \dot{\bm{s}}_k,\dot{\bm{s}}_{k+1}, \bm{f}_k,\bm{\tau}_{k}) = 0 \\
& \alpha \bigl(\bm{s}_k, \dot{\bm{s}}_k, \bm{f}_k, \bm{\tau}_k \bigr) = 0,  k=1...N \\ 
& \beta \bigl(\bm{s}_k, \dot{\bm{s}}_k, \bm{f}_k, \bm{\tau}_k\bigr) \leq 0, k=1...N 
\end{align}



\add{where $\bm{s}:=\left[\bm{p};\theta;\bm{q}\right]$ is the full state of the system at sample $t$ along the trajectory, $\bm{p}$ is the CoM position of the trunk in the world frame, $\theta$ is pitch angle, and $\bm{q}$ is the joint angle}. $J$ and $w$ are final and additive costs to jump to a particular height and distance while minimizing energy, $\delta t$ is the time between sample points $t$ and $t+1$, and $N$ is the total number of samples along the trajectory (i.e. $N = N_{dc} + N_{sc} + N_{fl}$). $\bm{f}$ is the force at the foot contact, and $\bm{\tau}$ is the joint torque.

The function $\alpha(\cdot)$ represents initial joint and body configurations, pre-landing configuration, and final body configuration. The function $\beta(\cdot)$ captures various constraints on the jumping motions, including joint angle/velocity/torque limits, friction cone limits, minimum ground reaction forces,  and geometric constraints related to the ground and obstacle clearance. 
The full-body dynamic constraint in the discrete form is represented by the function $\lambda (\cdot)$ as follows
\begin{equation}\label{eq:full_dynamics_2D}
    \begin{bmatrix}\bm{H} & -\bm{J}_c^T \\ -\bm{J}_c^T & \mathbf{0} \end{bmatrix} \begin{bmatrix}
    \bm{\ddot{s}} \\ \bm{f}
    \end{bmatrix}= \begin{bmatrix}-\bm{C}\bm{\dot s}-\bm{g} + \bm{S}\bm{\tau}+\bm{S}_{f}\bm{\tau}_{f} \\ \bm{\dot J}_{c}(\bm{s})\bm{\dot s}\end{bmatrix}, \nonumber
\end{equation}
where the mass matrix is represented by $\bm{H}$, the Coriolis  and centrifugal terms are represented by $\bm{C}$, and gravity vector is denoted as $\bm{g}$. $\bm{J}_c$ is the Jacobian expressed at the foot contact, $\bm{S}$ and $\bm{S}_{fric}$ are distribution matrices of actuator torques $\bm{\tau}$ and joint friction torques $\bm{\tau}_{fric}$. The dimensions of $\bm{J}_c$ and $\bm{f}$ are determined by the contact phases.
\add{Our designed ILC is a force-based controller. For the first trial, we utilize the optimal contact force obtained from trajectory optimization to initialize the learning process. The implementation of the low-level controller is explained in Appendix V.B}.

\subsection{Iterative Learning Control for Target Tracking} \label{sec:ILC_controller}
In the following, we present an ILC design to perform multiple challenging jumping tasks while enhancing target accuracy and ensuring safety during the learning process. Firstly, we present a model for ILC, followed by an optimization design with three stages to learn skills for a simple task. Finally, we propose a procedure to accomplish more challenging maneuvers from the simple task.

\subsubsection{Dynamic Jumping Model for ILC}\label{sec:ILC_model}
In quadruped robots, a single rigid body dynamics model is favorable to use because it can capture the dominant dynamical relationship between the ground reaction force and the body trajectory for agile maneuvers, while avoiding heavy computation related to a full-body dynamics model \cite{YanranDingTRO,GabrielICRA2021, Quad-SDK, park2017high}.
For a more detailed explanation of the simplified dynamics of quadruped jumping, please refer to the Appendix \ref{subsec:dynamics}. With our ILC approach, leg and contact dynamics can be considered as unknown dynamics introduced to the SRB dynamic model. Additionally, we consider other factors such as deformation of feet \cite{Zachary_RAL_online_calibration_2022} and non-constant friction coefficients as uncertainties. 

At future trial $k+1$, the robot state can be written recursively as a combination of initial configuration at the beginning $\bm{x}_{k+1,0}$ and control inputs $\bm{u}_{k+1,t}$ over the trial $k+1$.
\begin{equation}
        \bm{x}_{k+1,t+1} =\bm{A}^{t+1} \bm{x}_{k+1,0} + \sum_{j=0}^t \bm{A}^{t-j}\bm{B}_{k+1,j}\bm{u}_{k+1,j}
\end{equation}
In our problem, we consider the robot starts at the same initial condition for all trials, i.e., $\bm{x}_{k,0}$ is unchanged. Since $\bm{B}_{k+1,t}$ depends on future states, we can approximate to the its values after execute the trial $k$, denoted by $\bm{B}_{k,t}$, i.e. $\bm{B}_{k+1,t} \approx \bm{B}_{k,t}$. Therefore, given the input offset between two consecutive jumping as
\begin{equation}
    \Delta \bm{u}_{k,t} = \bm{u}_{k+1,t}-\bm{u}_{k,t},
\end{equation}
we can obtain the difference of robot states between two consecutive trials as
\begin{align}
\Delta \bm{x}_{k+1,t} &= \bm{x}_{k+1,t}- \bm{x}_{k,t} \nonumber \\ & \approx \sum_{j=0}^{t-1} \bm{A}^{t-1-j}\bm{B}_{k,j} \Delta \bm{u}_{k,j}
\end{align}
Due to the periodic reference, the error model of the next jumping $k+1$ can be estimated as: 
\begin{align}\label{eq:ILC_predict_error_kt}
    \bm{e}_{k+1,t} &=\bm{x}^{ref}_{k+1,t}-\bm{x}_{k+1,t}= \bm{e}_{k,t} - (\bm{x}_{k+1,t}- \bm{x}_{k,t})\nonumber\\
    & = \bm{e}_{k,t} - \sum_{j=0}^{t-1} \bm{A}^{t-1-j}\bm{B}_{k,j} \Delta \bm{u}_{k,j}
\end{align}


For jumping motion, there is no control input is applied during aerial phase. In other word, 
\begin{equation}
    \bm{u}_{k,t}=\bm{0}, \forall t> N_c = N_{dc} + N_{sc}
\end{equation}
Thus, we can rewrite the error model (\ref{eq:ILC_predict_error_kt}) as following:

\[\bm{e}_{k+1,t} = \left\{
  \begin{array}{lr}
    \bm{e}_{k,t} -  \sum_{j=0}^{t-1} \bm{A}^{t-1-j}\bm{B}_{k,j} \Delta \bm{u}_{k,j} : t \leq N_c\\
   \bm{e}_{k,t} -  \sum_{j=0}^{N_c-1} \bm{A}^{N_c-1-j}\bm{B}_{k,j} \Delta \bm{u}_{k,j} : t >N_c
  \end{array}
\right.
\]

Given the predicted error model, we now present an opti-
mization strategy with three stages for accurate target jumping.

\subsubsection{Our proposed ILC} \label{sec: ILC_optimization}
In the following, we first design a baseline ILC-MPC for quadruped jumping. Building upon this, we propose an ILC with multi-stage optimization to enable accurate target jumping with a long aerial phase. The baseline ILC-MPC aims to minimize the error of the entire trajectory and can be formulated as an optimization problem as follows:
\begin{subequations} \label{eq:baseline_ILC_MPC}
\begin{align}
\underset{\Delta \bm{u}_{k,t}}{\text{min}}  & \sum_{}  \textrm{ } \bm{e}_{k+1,t}^T \bm{Q}_{t}^e \bm{e}_{k+1,t}+ \sum_{t=1}^{N_{c}} \Delta \bm{u}_{k,t}^T \bm{Q}_{t}^u \Delta \bm{u}_{k,t}\nonumber\\
\text{s.t. }
             & \bm{\mathcal{T}}_{k,t} (\bm{u}_{k,t}+\Delta \bm{u}_{k,t}) \in \mathcal{I}^{mdc}_{k,t} \cap \mathcal{I}^{sat}_{k,t} , \\
            & \bm{u}_{k,t}+\Delta \bm{u}_{k,t} \in \mathcal{I}^{contact}_{k,t} \cap \mathcal{E}^{phases}_{k,t},
\end{align}
\end{subequations}
 where the transition matrice 
\begin{align}
    \bm{\mathcal{T}}_{k,t}=\bm{J}(\bm{q}_{k,t})^\top \bm{R}_{k,t}^\top
\end{align}
maps ground contact force (in the world frame) into the joint torque (in the body frame).
$\bm{Q}_t^e=diag(w_x, w_z, w_{\theta}, w_{\dot{x}}, w_{\dot{z}}, w_{\dot{\theta}})\in \mathbb{R}^6$ is weight matrix regarding to trajectory error term at time step $t$. $\bm{Q}_{t}^u \in \mathbb{R}^4$ is the weighting matrices to penalize the offset of control inputs between two consecutive trials.

Given the optimal offset $\Delta \bm{u}_{k,t}^*$ obtained from the baseline ILC-MPC design (\ref{eq:baseline_ILC_MPC}), the control input applied to the next trial $k+1$ is of the form:
\begin{equation}
     \bm{u}_{k+1,t}= \Delta \bm{u}_{k,t}^*+ \bm{u}_{k,t},
\end{equation}
where  $\bm{u}_{k,t}$ is the control input at trial $k$.

A challenge related to jumping is that we are only able to rely on the force control during the contact phases to accomplish the jumping task that involves long flight phase, while we need to ensure the safety during the whole jump. We have discovered that learning the entire trajectory as a baseline design (\ref{eq:baseline_ILC_MPC}) is not efficient because  it results in more bad failures at the first several trials, making it unsuitable for hardware transfer. Moreover, the data obtained from these failures might not even helpful, so the robot normally take more trials to complete the tasks.
To tackle this problem, we propose to separate more challenging tasks into different stages and master skills for each stage. Our proposed ILC is formulated as the following optimization problem:

\begin{subequations} \label{eq:optimize_ILC}
\begin{align}
\underset{\Delta \bm{u}_{k,t}}{\text{min}}  & \sum_{t\in S\lbrace i \rbrace}  \textrm{ } \bm{e}_{k+1,t}^T \bm{Q}_{t}^e \bm{e}_{k+1,t}+ \sum_{t=1}^{N_{c}} \Delta \bm{u}_{k,t}^T \bm{Q}_{t}^u \Delta \bm{u}_{k,t}\nonumber\\
\text{s.t. }
             & \bm{\mathcal{T}}_{k,t} (\bm{u}_{k,t}+\Delta \bm{u}_{k,t}) \in \mathcal{I}^{mdc}_{k,t} \cap \mathcal{I}^{sat}_{k,t} , \\
            & \bm{u}_{k,t}+\Delta \bm{u}_{k,t} \in \mathcal{I}^{contact}_{k,t} \cap \mathcal{E}^{phases}_{k,t},
\end{align}
\end{subequations}
The optimization stages $S\lbrace i \rbrace$ ($i = 1, 2, 3$) represent different timing intervals selected to optimize trajectory errors. \add{Stage I consists of initial trials, followed by Stage II for the next set of trials. Stage III continues until the robot reaches the target. These sequential stages are described in detail as follows.}
\begin{itemize}
    \item \textit{Stage I - Contact Priority} : In this stage, we only consider optimizing trajectory errors in all-leg contact phase and rear-leg contact phase. For this purpose, we set $\bm{Q}_{t}^e \neq \bm{0}, \forall t \in S\lbrace 1 \rbrace =\left [ 1, N_c \right]$; otherwise, $\bm{Q}_{t}^e = \bm{0}$. We only collect data from these contact phases for this learning stage.
    \item  \textit{Stage II - Hybrid-optimization} : We shift the optimization windows to the rear-leg contact phase and flight phase (i.e., $S\lbrace 2 \rbrace = \left [ N_{dc}, N\right]$), which aims to minimize the trajectory errors during these phases.
    We particularly set $\bm{Q}_{t}^e \neq 0, \forall t\in S\lbrace 2 \rbrace$, and $\bm{Q}_{t}^e= \bm{0}$ otherwise.
    \item \textit{Stage III - Goal Priority}:  This concluding stage aims to make the robot jump to the target at high accuracy. The term "\textit{goal-priority}" refers to the learning technique in which we use only the weight for trajectory errors $\bm{Q}_{t}^e$ at the end of the trajectory while setting all the other weights as $0$. In other words, $\bm{Q}_{t}^e \neq \bm{0}, t\in S\lbrace 3 \rbrace = N$; otherwise, $\bm{Q}_{t}^e = \bm{0}$.
    We repeat the learning process until the robot reaches the desired position target.
\end{itemize}

In the following, we present the constraints and fast-solving QP formulation for the optimization problem (\ref{eq:optimize_ILC}), then describe the low-level controller in more detail.

\textit{a) \underline{Constraints:}} The constraints enforced in the optimization formulation (\ref{eq:optimize_ILC}) are described in detail as follows
\begin{itemize}
    \item $\mathcal{I}^{mdc}_{k,t}$ denotes the set for the inequality constraints related to motor dynamic constraints presented in section \ref{sec: ILC_MDC_modeling}. These constraints aim to
represent the true system limits, playing an important role in
realizing the jumping motion in hardware. Since the joint torques $\bm{\tau}$ relate to the joint velocities $\dot{\bm{q}}$ and the applied voltage $\bm{V}$ as
\begin{align}
   \bm{\tau}_{k,t} =  \frac{\bm{V} - \sigma \dot{\bm{q}} }{\rho}, ~ \bm{V}  \in [-\bm{V}_{min}, \bm{V}_{max}],
\end{align}
these constraints can be rewritten as
\begin{align}
    \mathcal{I}^{mdc}_{k,t}=\left [ \frac{\bm{V}_{min} - \sigma \dot{\bm{q}}_{k,t} }{\rho}, \frac{\bm{V}_{max} - \sigma \dot{\bm{q}}_{k,t} }{\rho} \right]
\end{align}



\item The set $\mathcal{I}^{sat}_{k,t}=[-\bm{\tau}_{max},\bm{\tau}_{max}]$ represents the torque limits that can generate by the robot's motors.
    
\item $\mathcal{E}^{phases}_{k,t}$ represents for equality constraints that nullifies the force exerted on the swing legs defined by contact schedule. 

\item $\mathcal{I}^{contact}_{k,t}$ is a set of forces $f = [f_x, f_z]$ which satisfies inequality constraints on scheduled stance legs, i.e., force limits, and linearized friction cone to prevent slippery:
\[\mathcal{I}^{contact}_{k,t} \in \left\{
  \begin{array}{lr}
    0 < f_{min} \leq f_{z} \leq f_{max} \\
    |f_x| \leq \mu f_z
  \end{array}
\right.
\]
\end{itemize}


\textit{b) \underline{Fast-solving QP formulation}}:
To solve the optimization (\ref{eq:optimize_ILC}), we propose to formulate it as a Quadratic Programming for computational efficiency. Depending on Stage $S\lbrace i \rbrace$, we define a concatenation of actual errors along this window after executing trial $k$ as $\bm{e}_{k}\lbrace i \rbrace = [\bm{e}_{k,t_1}^T, \bm{e}_{k,t_2}^T, ..., \bm{e}_{k,t_n}^T]^T$, in which $t_1,..., t_n \in S \lbrace i \rbrace$. 
According to the error model in Section \ref{sec:ILC_model}, one can obtain the predicted trajectory errors along this contact-switch windows in a concatenated form as
\begin{equation}\label{eq:ILC_predict_error_k}
    \bm{e}_{k+1} \lbrace i \rbrace=\bm{e}_k \lbrace i \rbrace-\bm{G}_k \lbrace i \rbrace \Delta\bm{u}_{k},
\end{equation}
where the control offset $\Delta\bm{u}_{k}= \bm{u}_{k+1} - \bm{u}_{k} \triangleq [\Delta \bm{u}_{k,0}^T, \Delta \bm{u}_{k,1}^T, ..., \Delta \bm{u}_{k,N_c-1}^T]^T$.
$\bm{G}_k $ is a block matrix updated after each trial $k$. $\bm{G}_k \lbrace i \rbrace$ is derived from $\bm{G}_k $ by selecting the rows correspondingly with each Stage. For example, $\bm{G}_k \lbrace 1 \rbrace$ consists the row $1$ to $N_c$ of the block matrix $\bm{G}_k $, which is described in \textit{Stage I} of Optimization Schedule. The matrix $\bm{G}_k $ is of the following form:


\begin{align}
\bm{G}_k =\left[\begin{array}{cccc}
\bm{G}^k_{1,1} & \bm{G}^k_{1,2} & \cdots & \bm{G}^k_{1,N_c}\\
\bm{G}^k_{2,1} & \bm{G}^k_{2,2} & \cdots & \bm{G}^k_{2,N_c} \\
\vdots & \vdots & \ddots & \vdots\\
\bm{G}^k_{N_c,1} & \bm{G}^k_{N_c,2} & \cdots & \bm{G}^k_{N_c,N_c}\\
\hline
\bm{G}^k_{N_c+1,1} & \bm{G}^k_{N_c+1,2} & \cdots & \bm{G}^k_{N_c+1,N_c}\\
\vdots & \vdots & \ddots & \vdots\\
\bm{G}^k_{N,1} & \bm{G}^k_{N,2} & \cdots & \bm{G}^k_{N,N_c}\\
\end{array}\right], \nonumber
\end{align}
where its elements are defined as
\[\bm{G}^k_{m,n} = \left\{
  \begin{array}{lr}
     \bm{B}_{k,m-1} : & \textit{if} ~ m=n ~\textit{and} ~ m \leq N_c\\
    \bm{A}^{m-1} \bm{B}_{k,n} : & \textit{if} ~ m >n \\
    \bm{0}  & \textit{otherwise}
  \end{array}
\right.
\]
for $m\in \left[1,N \right]$ and $n \in \left[1, N_c \right]$.
We then construct the weighted matrices $\bm{Q}^{e}$ for the output errors and $\bm{Q}^{u}$ for the control input offsets in the concatenated form as following
\begin{equation}
\bm{Q}^{e} \lbrace i \rbrace= blkdiag\Big(\bm{Q}_{t_1}^e,\bm{Q}_{t_2}^e, ..., \bm{Q}_{t_n}^e  \Big); t_1,..., t_n \in S \lbrace i \rbrace \nonumber
\end{equation}
\begin{equation}
\bm{Q}^{u} \lbrace i \rbrace= blkdiag\Big( \underbrace{\bm{Q}_{1}^u..,\bm{Q}_{dc}^u}_{dc ~phase},\underbrace{\bm{Q}_{dc+1}^u..,\bm{Q}_{N_c}^u}_{sc~phase} \Big) \nonumber
\end{equation}
By defining,
\begin{subequations}
\begin{align}
    \bm{W}_k \lbrace i \rbrace & =2\Big(\bm{G}_{k-1}^{\top} \lbrace i \rbrace  \bm{Q}^e \lbrace i \rbrace  \bm{G}_{k-1} \lbrace i \rbrace +\bm{Q}^u \lbrace i \rbrace\Big)\\
    \bm{h}_k & =-2 \bm{G}_{k-1}^{\top} \lbrace i \rbrace  \bm{Q}^e \lbrace i \rbrace \bm{e}_{k-1} \lbrace i \rbrace 
\end{align}
\end{subequations}
we can formulate the constrained optimization problem (\ref{eq:optimize_ILC}) as a compact QP as follows
\begin{subequations} \label{eq: QP_ILC}
\begin{align}
&\underset{\Delta \bm{u}_{k}}{\text{minimize}} \textrm{ } \Big( \frac{1}{2}\Delta \bm{u}_{k}^T \bm{W}_k \Delta \bm{u}_{k} +  \Delta \bm{u}_{k}^T \bm{h}_k\Big)\nonumber\\
\text{s.t. }
            &\bm{\Psi}_k^{mdc} \Delta \bm{u}_k \leq \overline{\bm{\mathcal{I}}}_k^{mdc} (V_{max}, \dot{\bm{q}}_{k},\bm{u}_k, \bm{\mathcal{T}}_{k} )  \\
             & \bm{\Psi}_k^{mdc} \Delta \bm{u}_k \geq \underline{\bm{\mathcal{I}}}_k^{mdc} (V_{min}, \dot{\bm{q}}_{k},\bm{u}_k, \bm{\mathcal{T}}_{k} ) \\
             & \bm{\Psi}_k^{sat} \Delta \bm{u}_k \leq \overline{\bm{\mathcal{I}}}_k^{sat}(\bm{\tau}_{max},\bm{u}_k, \bm{\mathcal{T}}_{k})  \\
             &  \bm{\Psi}_k^{sat} \Delta \bm{u}_k \geq \underline{\bm{\mathcal{I}}}_k^{sat} (\bm{\tau}_{max},\bm{u}_k, \bm{\mathcal{T}}_{k})  \\
             & \bm{\Psi}_k^{cot} \Delta \bm{u}_k \leq \overline{\bm{\mathcal{I}}}_k^{contact} (f_{max}, \bm{u}_k, \bm{\mathcal{T}}_{k})  \\
             &  \bm{\Psi}_k^{cot} \Delta \bm{u}_k \geq \underline{\bm{\mathcal{I}}}_k^{contact} (f_{min}, \bm{u}_k, \bm{\mathcal{T}}_{k}) \\
            & \bm{\Psi}_k^{phase} \Delta \bm{u}_k = \bm{\mathcal{E}}^{phase}_{k}(\bm{u}_k)
\end{align}
\end{subequations}

where the matrices $\overline{\bm{\mathcal{I}}}_k^{mdc}$ and $\underline{\bm{\mathcal{I}}}_k^{mdc}$ represent upper and lower bounds on motor dynamic constraints, while $\overline{\bm{\mathcal{I}}}_k^{sat}$ and $\underline{\bm{\mathcal{I}}}_k^{sat}$ denote torque saturation limits. Additionally, $\overline{\bm{\mathcal{I}}}_k^{contact}$ and $\underline{\bm{\mathcal{I}}}_k^{contact}$ represent control force limits and the friction cone on stance legs. The term $\bm{\mathcal{E}}^{phase}_{k}$ is defined based on the contact schedule, nullifying the control force on the swing leg. 

\begin{figure*}[!t]
	\centering
	\subfloat[Distance errors $e_x \triangleq x_{T}-x_{target}$]{\centering
		\resizebox{0.33\linewidth}{!}{\includegraphics[trim={0cm 0cm 0cm 0cm},clip]{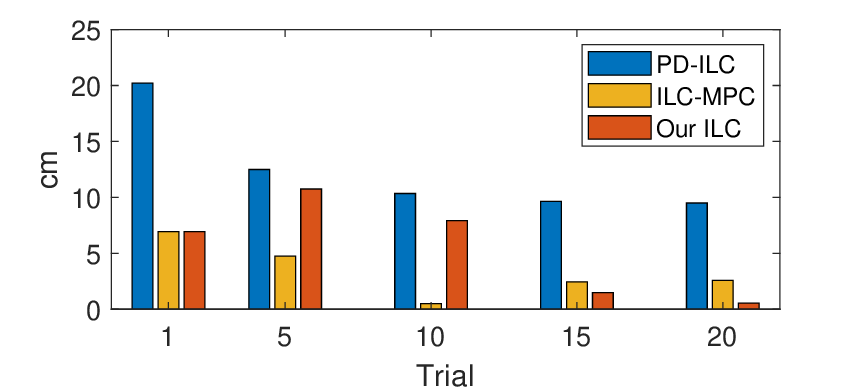}}\label{fig:softground_bar_x}
	}
	\subfloat[Height errors $e_y \triangleq y_{T}-y_{target}$]{\centering
		\resizebox{0.33\linewidth}{!}{\includegraphics[trim={0cm 0cm 0cm 0cm},clip]{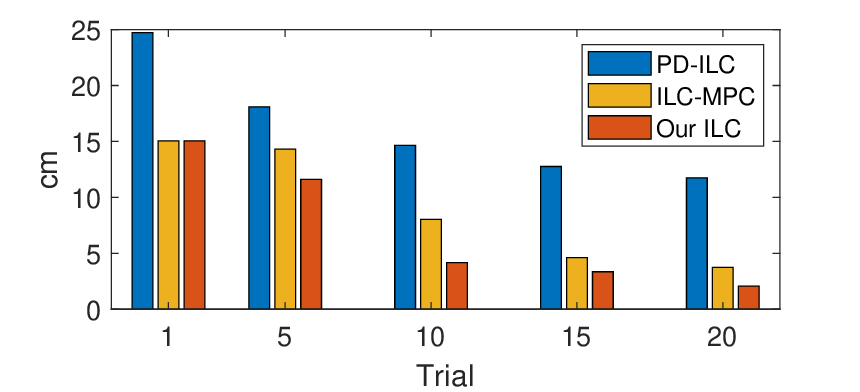}}\label{fig:softground_bar_y}
	}
	\subfloat[Orientation errors $ e_{\theta} \triangleq \theta_{T} - \theta_{target}$]{\centering
		\resizebox{0.33\linewidth}{!}{\includegraphics[trim={0cm 0cm 0cm 0cm},clip]{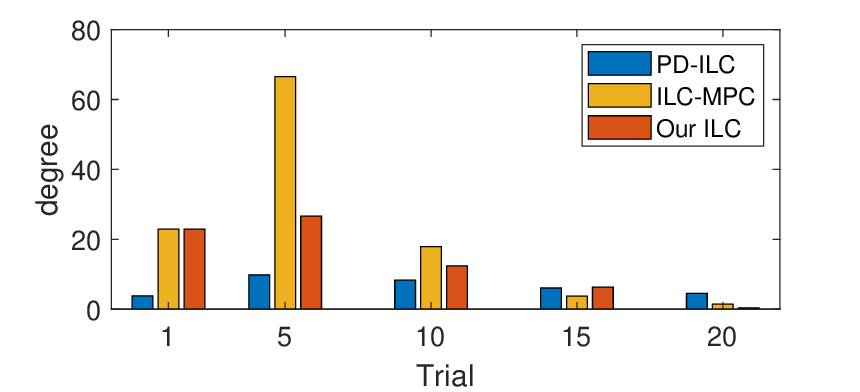}}\label{fig:softground_bar_pitch}
	}\\
	\caption{\textbf{Jump from soft ground}. The figures show the comparison between model-free PD-type ILC, baseline ILC-MPC, and our proposed ILC in terms of body trajectories and body orientation for the jumping forward to a target at $0.6~m$ from soft ground with $(K_p^G, K_d^G)=(2.10^3, 5.10^2)$.} 
	\label{fig:compare_jump_soft_ground}
\end{figure*}
\begin{figure*}[!t]
	\centering
	\subfloat[Distance errors $e_x \triangleq x_{T}-x_{target}$]{\centering
		\resizebox{0.33\linewidth}{!}{\includegraphics[trim={0cm 0cm 0cm 0cm},clip]{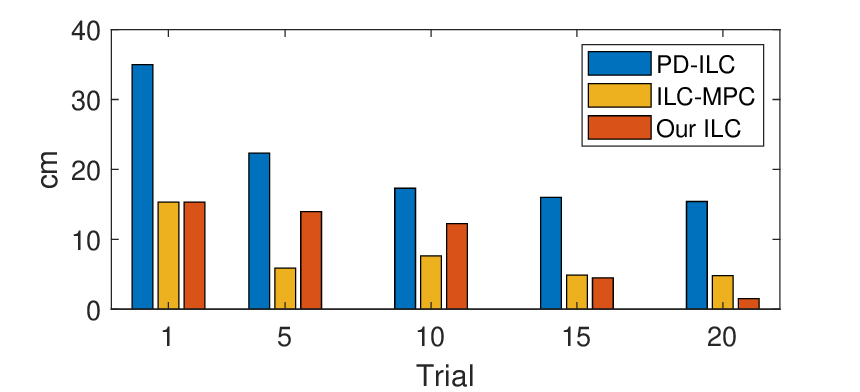}}\label{fig:hardground_bar_x}
	}
	\subfloat[Height errors $e_y \triangleq y_{T}-y_{target}$]{\centering
		\resizebox{0.33\linewidth}{!}{\includegraphics[trim={0cm 0cm 0cm 0cm},clip]{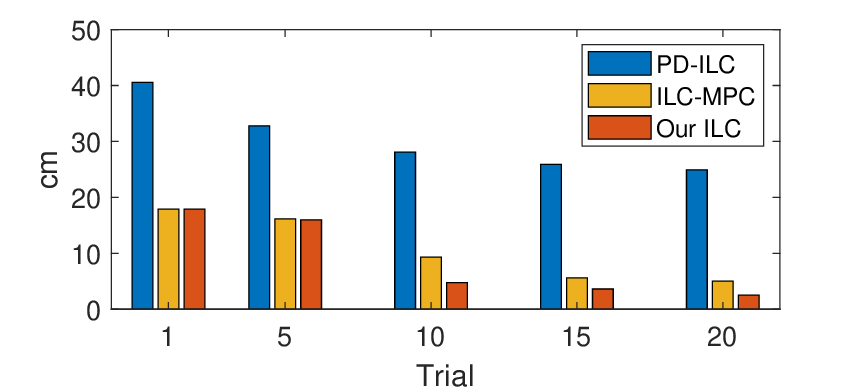}}\label{fig:hardground_bar_y}
	}
	\subfloat[Orientation errors $e_{\theta} \triangleq \theta_{T} - \theta_{target}$]{\centering
		\resizebox{0.33\linewidth}{!}{\includegraphics[trim={0cm 0cm 0cm 0cm},clip]{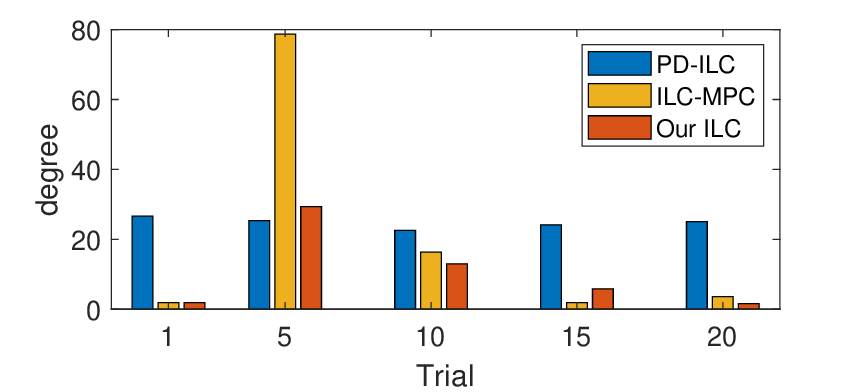}}\label{fig:hardground_bar_pitch}
	}\\
	\caption{\textbf{Jump from hard ground}. The plots show the learning progress over trials with model-free PD-type ILC, baseline ILC-MPC, and our proposed ILC when the robot jumping from the hard ground with $(K_p^G, K_d^G)=(2.10^4, 3.10^3)$ .} 
	\label{fig:compare_jump_hard_ground}
\end{figure*}
Having presented the ILC for a simple jump, we then propose a procedure to apply what the robot learns from the simple task to accomplish more challenging tasks.
\subsubsection{Learning more challenging tasks from simple practices}\label{sec:simple_to_complex}

The ILC design for this generalization task is illustrated in detail in Fig. \ref{fig:simple_to_complex_framework}.
We start with a successful simple task, e.g., a short jump that has already been learned by following three-staged optimization mentioned earlier in Section \ref{sec: ILC_optimization}. By leveraging the learning result from the simple task (i.e. optimal control force $\bm{u}_s^*$), we enable the robot to complete new goals much faster in comparison with learning from scratch. To this end, we propose to evaluate only the actual final distance and orientation. As formulated earlier in Section \ref{sec:ILC_model}, given the actual errors $\bm{e}_{k,N}$ at the end of the flight phase of the current trial $k$, the predicted errors at the end of the next trial can be computed as follows:
\begin{align}
    \bm{e}_{k+1,N}= \bm{e}_{k,N} -  \sum_{j=0}^{N_c-1} \bm{A}^{N_c-1-j}\bm{B}_{k,j} \Delta \bm{u}_{k,j},
\end{align}
Then, for each trial, we propose to solve the following \textit{goal-priority} optimization problem with constraints as follows: 
\begin{subequations} 
\begin{align}
\underset{\Delta \bm{u}_{k,t}}{\text{min}} \textrm{ }  &  \bm{e}_{k+1,N}^T \bm{Q}_{N}^e \bm{e}_{k+1,N}+ \sum_{t=1}^{N_{c}}   \Delta \bm{u}_{k,t}^T \bm{Q}_{t}^u \Delta \bm{u}_{k,t}\nonumber\\
\text{s.t. }
             & \bm{\mathcal{T}}_{k,t} (\bm{u}_{k,t}+\Delta \bm{u}_{k,t}) \in \mathcal{I}^{sat}_{k,t} \cap \mathcal{I}^{mdc}_{k,t} \\
            & \bm{u}_{k,t}+\Delta \bm{u}_{k,t} \in \mathcal{I}^{contact}_{k,t} \cap \mathcal{E}^{phase}_{k,t}
\end{align}
\end{subequations}\label{eq:ILC_with_prior_experience}
The iteration procedure is stopped until the robot reaches the desired position target. To solve this problem, we formulate it as Quadratic Programming. This is similar to Stage 3 in the optimization (\ref{eq: QP_ILC}), thus consequently omitting the details here. In the low-level controller, we can persist in utilizing the joint reference profile ($\bm{q}_d, \dot{\bm{q}}_d$) designed for a simple jump. 
Since our approach relaxes the emphasis on joint tracking performance to focus on the target reaching. We can utilize the joint reference obtained from a simple jump for more challenging tasks. 

\begin{figure*}[!t]
	\centering
	\subfloat[Distance errors $e_x \triangleq x_{T}-x_{target}$]{\centering
		\resizebox{0.33\linewidth}{!}{\includegraphics[trim={0cm 0cm 0cm 0cm},clip]{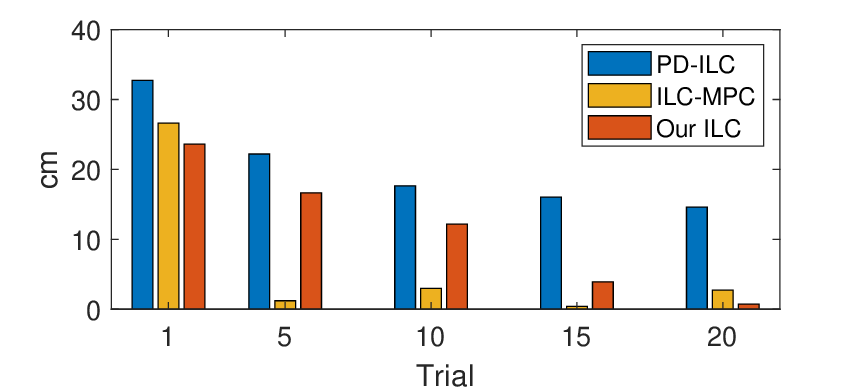}}\label{fig:mass_2kg_bar_x}
	}
	\subfloat[Height errors $e_y \triangleq y_{T}-y_{target}$]{\centering
		\resizebox{0.33\linewidth}{!}{\includegraphics[trim={0cm 0cm 0cm 0cm},clip]{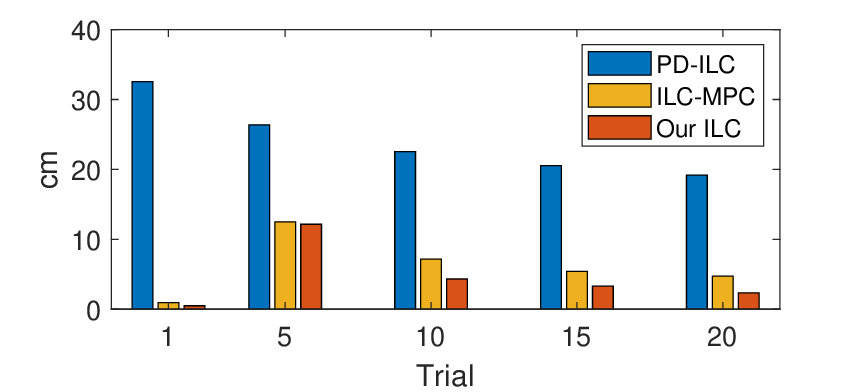}}\label{fig:mass_2kg_bar_y}
	}
	\subfloat[Orientation errors $e_{\theta} \triangleq \theta_{T} - \theta_{target}$]{\centering
		\resizebox{0.33\linewidth}{!}{\includegraphics[trim={0cm 0cm 0cm 0cm},clip]{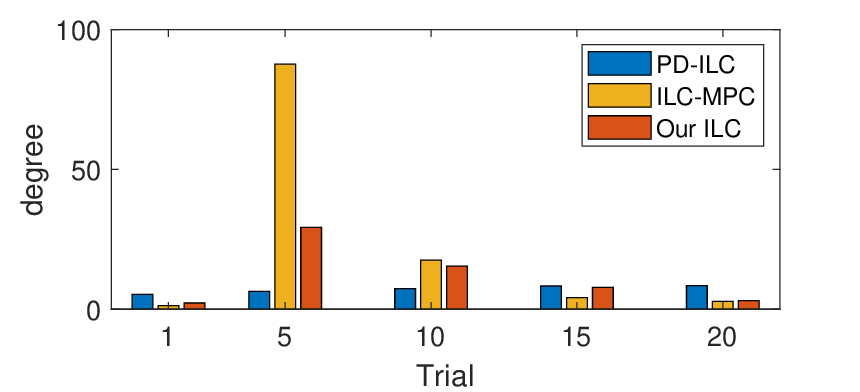}}\label{fig:mass_2kg_bar_pitch}
	}\\
	\caption{\textbf{Jump while carrying an unknown payload}. The figures show body trajectories and orientation over learning trials for jumping to target  $60~cm$ while carrying unknown load of $2~kg$. Our proposed ILC approach can enable robot jumping to the target accurately.} 
	\label{fig:carrying_mass_compare}
\end{figure*}


\section{Results}
\label{sec:ILC_Results}
This section presents the validation of our framework and the comparison with recent methods for target jumping in simulation and hardware experiments.

\begin{table*}
  \begin{minipage}{1\linewidth}
    \centering
    \begin{tabular}{ *{9}{c} }
      \toprule
      \makecell{} & \makecell{Approaches} & \makecell{trial 1 \\ $\lbrace e_{x}, e_{y}, e_{\theta} \rbrace$ } & \makecell{trial 5 \\ $\lbrace e_{x}, e_{y}, e_{\theta} \rbrace$} & \makecell{trial 10\\ $\lbrace e_{x}, e_{y}, e_{\theta} \rbrace$} & \makecell{trial 15 \\ $\lbrace e_{x}, e_{y}, e_{\theta} \rbrace$} & \makecell{trial 20 \\ $\lbrace e_{x}, e_{y}, e_{\theta} \rbrace$} & \makecell{Number of \\ bad landing angle} & \makecell{Target \\ reaching?} \\
      \midrule
      \midrule
     & PD-type ILC & 20, 25, 3 & 13, 18, 12 & 10, 15, 8 & 9, 13, 5 & 9, 12, 3 & 1 & No\\
     Soft Ground & ILC-MPC & 7, 15, 23 & 5, 15, 70 & 1, 8, 18 & 3, 5, 4 & 3, 4, 2 & 3 & No \\
     & Our ILC & 7, 15, 23 & 10, 12, 25 & 8, 4, 8 & 2, 3, 4 & \textbf{0.5, 2, 0.5} & \textbf{0} & \textbf{Yes} \\
      \midrule
     & PD-type ILC & 35, 40, 25 & 23, 33, 24 & 18, 28, 20 & 16, 26, 20 & 16, 24, 20 & 1 & No\\
     Hard Ground & ILC-MPC & 16, 18, 3 & 6, 16, 80 & 8, 10, 17 & 5, 5, 3 & 5, 4, 4 & 4 & No \\
     & Our ILC & 16, 18, 3 & 14, 16, 26 & 13, 5, 15 & 4, 3, 4 & \textbf{1, 2, 1} & \textbf{0} & \textbf{Yes} \\
     \midrule
     & PD-type ILC & 35, 35, 5 & 22, 28, 6 & 18, 22, 7 & 17, 20, 7 & 16, 20, 7 & 1 & No\\
     Unknown Weight & ILC-MPC & 25, 2, 2 & 2, 12, 90 & 3, 7, 15 & 0.5, 4, 5& 2, 3, 1.5 & 6 & No \\
     ($2~kg$)& Our ILC & 23, 1, 2 & 18, 12, 26 & 12, 5, 10 & 3, 2, 1.5 & \textbf{0.5, 2, 1.5} & \textbf{0} & \textbf{Yes} \\
      \bottomrule
    \end{tabular}
    \caption{Comparison between ILC approaches when performing a single jumping task with various ground and model uncertainties. $\lbrace e_x, e_y, e_{\theta}\rbrace (cm, cm, ^0)$ represent the final landing locations error and landing angle error. Our approach allows for accurate target jumping after only $20$ trials, while also ensuring a small landing angle. A landing angle greater than $70$ degrees is considered a bad landing angle, as it becomes difficult for advanced landing controllers to safely recover the robot in such instances \cite{quad_landing_nn, QuannICRA19,reactive_landing}}\label{tab:comparison_ILCs}
  \end{minipage}%
\end{table*}



 \begin{figure}[!t]
    	\centering
    	{\centering
    		\resizebox{1\linewidth}{!}{\includegraphics[trim={0cm 0cm 0cm 0cm},clip]{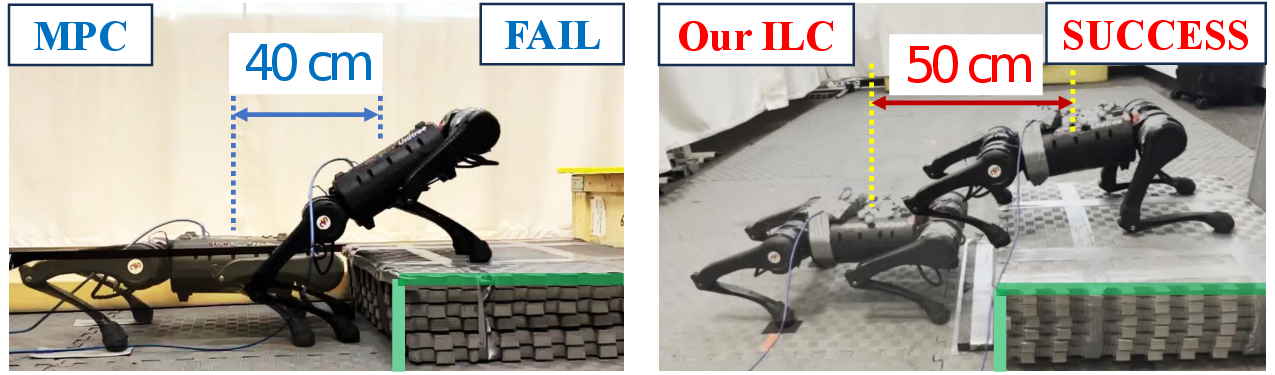}}}\\
    	\caption{\textbf{MPC vs Our ILC.} Robot performs jumping on box at $(x,z)=(50,20)cm$ with MPC (left) and our proposed ILC (right)} 
    	\label{fig:mpc_and_ilc}
\end{figure}

\begin{figure*}[!t]
	\centering
	{\centering
		\resizebox{0.9\linewidth}{!}{\includegraphics[trim={0cm 1cm 0cm 0.5cm},clip]{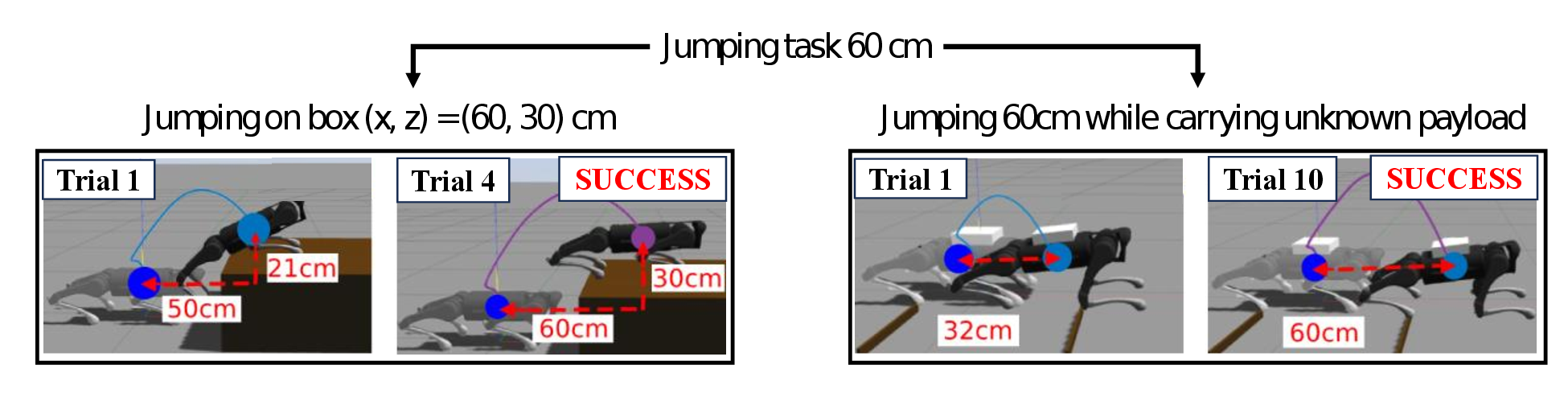}}
	}
        {\centering
		\resizebox{0.46\linewidth}{!}{\includegraphics[trim={0cm 0cm 0cm 0cm},clip]{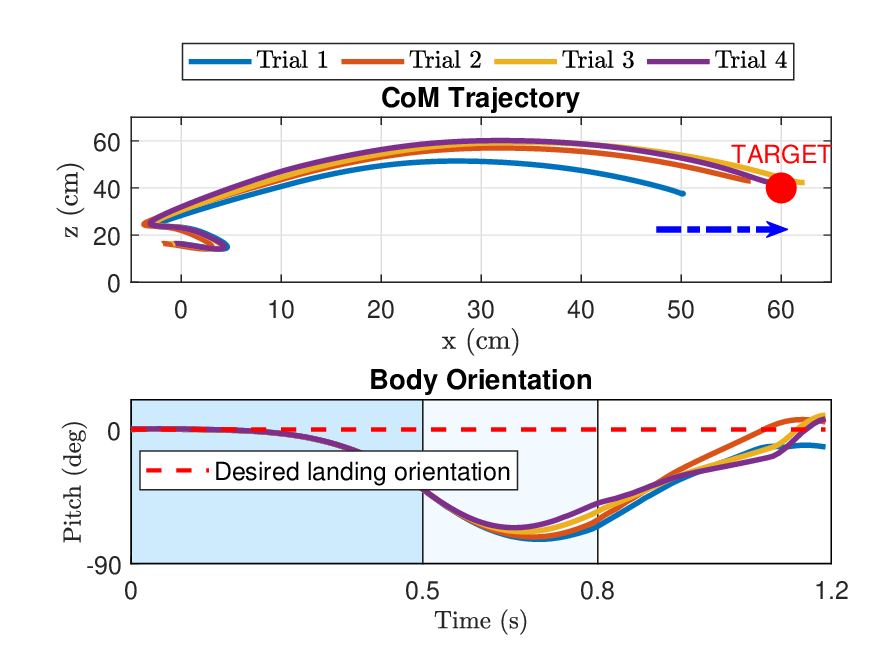}}
  		\resizebox{0.46\linewidth}{!}{\includegraphics[trim={0cm 0cm 0cm 0cm},clip]{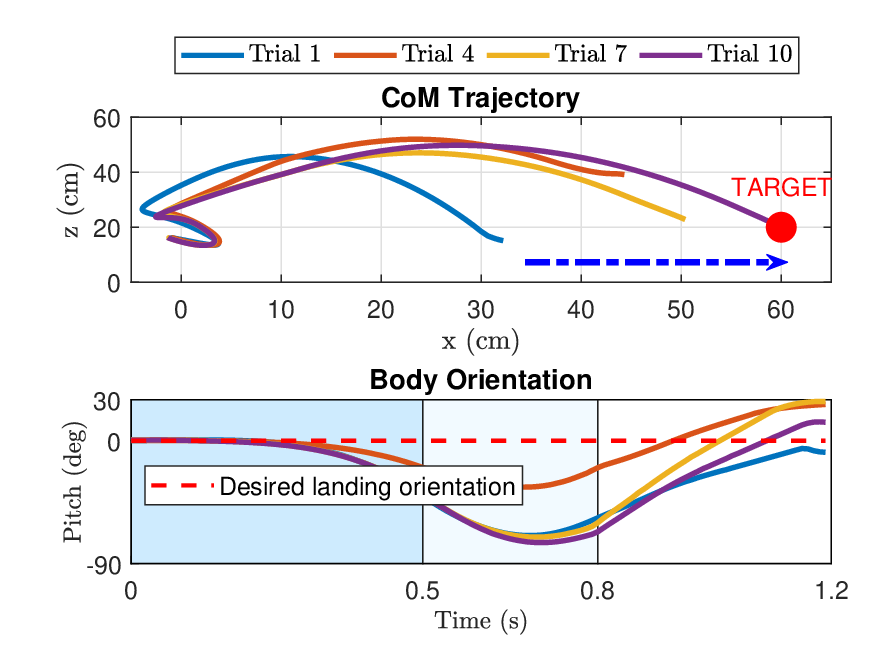}}
	}
	
	\caption{\textbf{Simulation: Learning to accomplish more challenging tasks from a simpler task}. The left figures show the learning process for box jumping after the robot masters its jumping skills for $60cm$ on the ground. Initially, the robot fails to jump on the box, but after three attempts, it successfully jumps on the box. The dark blue and light blue areas in the figures represent the all-legs contact and rear-legs contact phase, respectively. The right figures describe the process of learning how to jump with an unknown payload to reach a target of $60~cm$. The mass of the load is $2~kg$, which is about $20\%$ body weight. This introduces unknown additional weight and momentum to the robot model, making the learning process more challenging.}
	\label{fig:jump_simulation}
\end{figure*}

\subsection{Comparative Analysis} 
We verify the proposed framework with the Unitree A1 robot \cite{unitreeA1}. In the following, we compare our proposed ILC with other ILC and MPC approaches for target jumping on quadruped robots. For these comparisons, we focused on highly dynamic jumping motions with an aerial period of $400 ms$, which is sufficiently long to validate target jumping accuracy.

\subsubsection{Comparison with other ILC approaches}
PD-type ILC has recently been shown effectively to enable accurate pronking motions with short aerial phases \cite{pronking_ILC}.
Thus, we select and implement a PD-type ILC from \cite{pronking_ILC} to perform jumping maneuvers and validate its efficiency for target jumping. \add{We further compare with our designed baseline ILC-MPC, as formulated in the optimization (\ref{eq:baseline_ILC_MPC}). The comparisons are validated for quadruped jumping with a long flight phase under various uncertainties (e.g., ground contact coefficients) and unknown disturbance (e.g., unknown load).}
\textit{a)\underline{ Ground uncertainties}}: We consider two types of ground, i.e. hard and soft contact ground, which has stiffness and damping coefficients as $(K_p^G, K_d^G)=(2.10^4, 3.10^3)$ and $(K_p^G, K_d^G)=(2.10^3, 5.10^2)$, respectively. These ground contact parameters are unknown to both learning controllers. We also utilize the same trajectory optimization reference, derived in Section \ref{sec: ILC_TO} for the $60~cm$ forward jumping. To evaluate joint tracking performance at each trial $k$, we define the criterion as the average of joint angle error over the whole jump as $\epsilon_k =\sum_{t=1}^{t=N} \frac{|q_k(t) - q_{ref}(t)|}{N}$.
The evaluation for target tracking is based on the actual position and orientation of the robot at $t=N$.

\textit{PD-type ILC:}  It significantly reduces the joint tracking errors over trials regardless of ground contact properties. 
The tracking errors on rear thigh and rear calf motors reduce about $62\%$ after $20$ trials. 
However, the PD-type ILC learning is unable to drive the robot to reach the target accurately. The robot trajectory converges to a final location that is far short of $10-15cm$ to the target after $20$ trials, as shown in Fig. \ref{fig:compare_jump_soft_ground}a and Fig. \ref{fig:compare_jump_hard_ground}a. Also, there still exists a deviation in the robot landing angle from its reference at the end of the trajectory, as illustrated in Fig. \ref{fig:compare_jump_soft_ground}c and Fig. \ref{fig:compare_jump_hard_ground}c.

\NEW{\textit{Baseline ILC-MPC:} Compared to the PD-type ILC, the baseline ILC-MPC approach significantly reduces the position target error. After 20 trials, the robot reaches closely to the target with the position errors of $3cm$ and $5cm$ for soft ground and hard ground, as shown in Fig.\ref{fig:compare_jump_hard_ground}a and (Fig.\ref{fig:compare_jump_soft_ground}a respectively. However, we observe bad landing poses with landing angles about $70^0-80^0$ at trial $\lbrace 4,5,6 \rbrace $ for the soft ground case, and trial $\lbrace 4,5,6,7 \rbrace $ for the hard ground case. This bad landing leads to failure to recover the robot safely at the end. 
}

\textit{Our ILC method:} We divide the learning process into 3 stages. Stage I consists of the first five trials, followed by stage II which has five trials. Stage III will continue until the robot reaches the target. Our proposed approach outperforms the PD-type ILC and the baseline ILC-MPC in both position and pose target accuracy. Difference from PD-type ILC, our method allows the robot to focus on the final target jumping result, not just the joint angle trajectory, which does not present well the jumping performance.
In other words, the robot learns to sacrifice a certain level of whole trajectory tracking accuracy (e.g. joint profile) for only desired targets. 
Despite the ground uncertainties, our method enables the robot to jump to the target after $15-20$ trials (see Fig. \ref{fig:compare_jump_soft_ground}a and Fig. \ref{fig:compare_jump_hard_ground}a) with minimal deviations of $1cm$ in distance and $2cm$ in height, respectively. As shown in Fig. \ref{fig:compare_jump_soft_ground}c and Fig. \ref{fig:compare_jump_hard_ground}c, after $20$ trials, the robot reaches the target with the error of landing angle only $0.5^0-1^0$, regardless of unknown ground contact coefficients. 

We can observe that the landing angle can reach $25^0$ at the $5^{th}$ trial with the proposed ILC. This can be explained by the fact that we prioritize optimizing the contact phases instead of the final target during the first 5 trials. However, it is worth mentioning that $25^0$ is considered a normal landing angle, and many existing landing controllers are proven to handle well this landing posture ~\cite{quad_landing_nn,falling_cat,QuannICRA19,reactive_landing,landing_tail,chuongjump3D}.

\textit{b) \underline{Unknown Disturbances}}:
We evaluate the performance of these learning controllers with unknown disturbance, e.g. the robot carries an unknown load mass. The disturbance adds an unknown inertial and mass to the robot model. The mass weight is $2~kg$.
As we can see in Fig. \ref{fig:carrying_mass_compare}a, Fig. \ref{fig:carrying_mass_compare}b, and Fig. \ref{fig:carrying_mass_compare}c, the proposed ILC outperforms the other ILC methods. 
With PD-type ILC, we can still observe some large errors of $15-20cm$ in distance and height after $20$ trials. ILC-MPC fails to enable the landing safely at trial 5, with the landing angle around $90^0$.  In contrast, with our approach, the robot leaps to the target accurately with a distance error of $<1cm$ only, and the robot consistently maintains
small landing angles (e.g. less than $20^0$) during the
the whole learning process.

\NEW{The Table \ref{tab:comparison_ILCs} provides a summary of target errors at some trials, number of failures and success for all ILC approaches under various uncertainties. This table is associated with Fig. \ref{fig:compare_jump_soft_ground}, \ref{fig:compare_jump_hard_ground}, and \ref{fig:carrying_mass_compare}.
For all uncertainties (e.g ground contact coefficients) and
unknown disturbance (e.g. unknown load) presented, our approach requires fewer iterations to reach closely the position target, while achieving smaller errors in the pose target.} This can be intuitively explained for several reasons. The first reason is the effectiveness of the proposed optimization schedule, which enables us to leverage muscle memory for enhanced learning speed and safety. Secondly, the effectiveness of the SRB model allows us to integrate it into the rigorously formulated optimization problem, seeking optimal control actions efficiently.

\subsubsection{Our proposed ILC and recent MPC}


\begin{figure}[!t]
	\centering
	{\centering
		\resizebox{0.9\linewidth}{!}{\includegraphics[trim={0cm 0cm 0cm 0cm},clip]{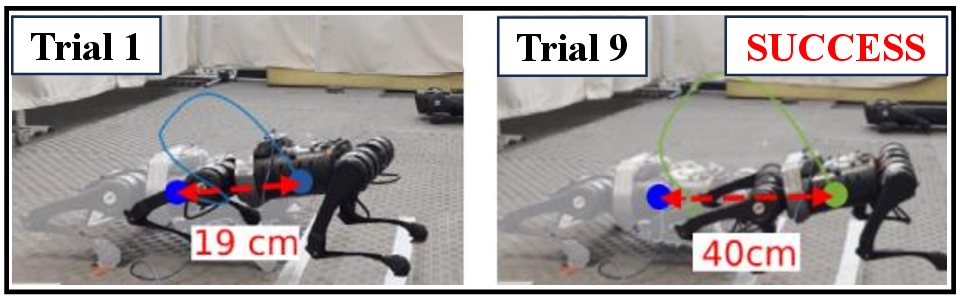}}
	}
    {\centering
		\resizebox{1\linewidth}{!}{\includegraphics[trim={0cm 0.3cm 0cm 0cm},clip]{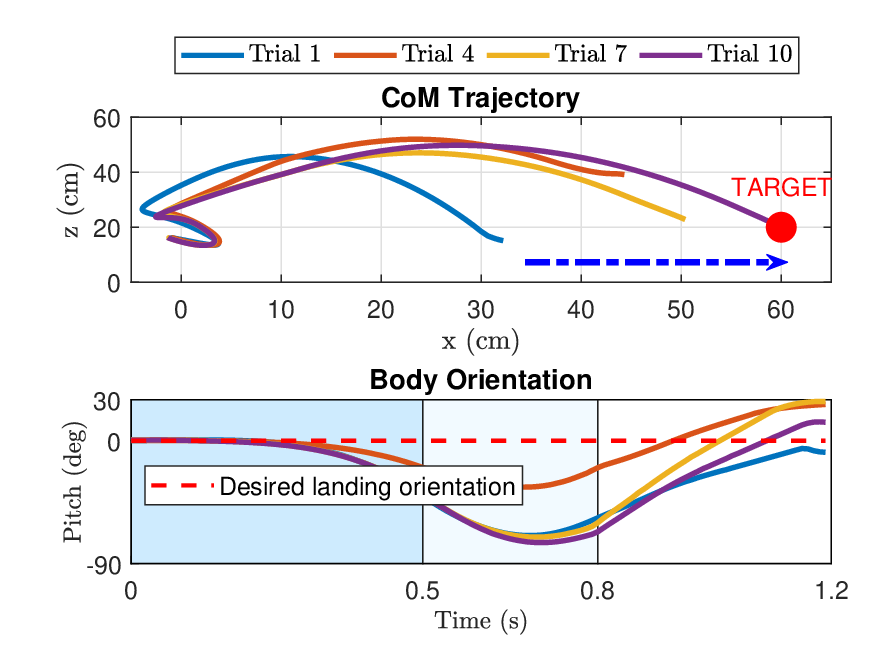}}
	}
	
	\caption{\textbf{Baseline task - jumping forward 40cm.} The figures show the learning progress of forward jumping $40cm$ in hardware. Instead of directly learning in hardware, the task was first learned in simulation. After practice necessary skills in simulation, the learning was transferred to the robot hardware. The robot only required $9$ additional trials to achieve the target, with the landing angle error being only a few degrees.
 }
	\label{fig:jump_40d0h_traj}
\end{figure}

\begin{figure*}[!t]
	\centering
	{\centering
		\resizebox{0.9\linewidth}{!}{\includegraphics[trim={0cm 0cm 0cm 0cm},clip]{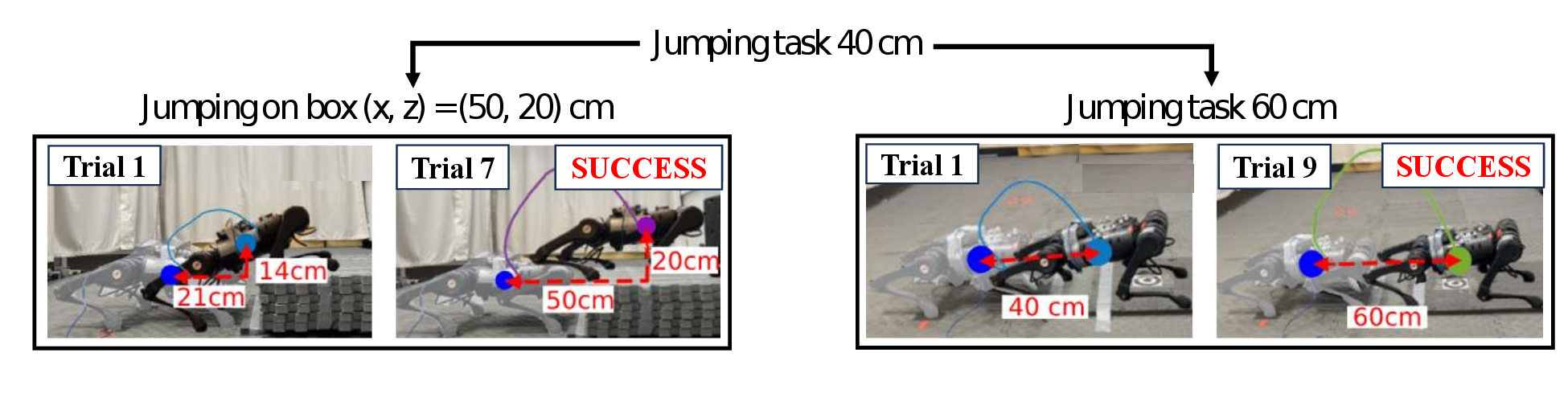}}
	}
        {\centering
          \resizebox{0.45\linewidth}{!}{\includegraphics[trim={0cm 0cm 0cm 0cm},clip]{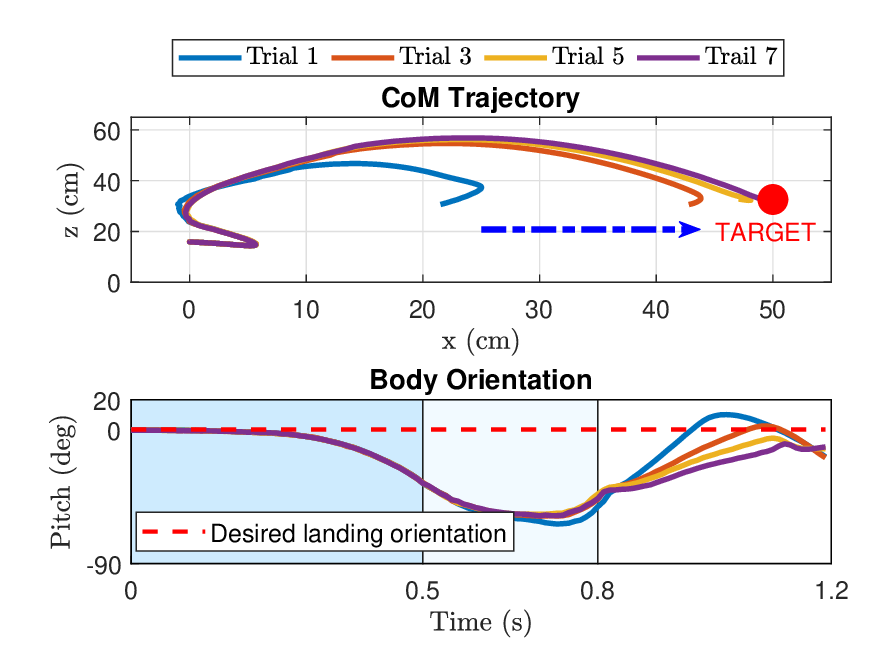}}
	   \resizebox{0.45\linewidth}{!}{\includegraphics[trim={0cm 0cm 0cm 0cm},clip]{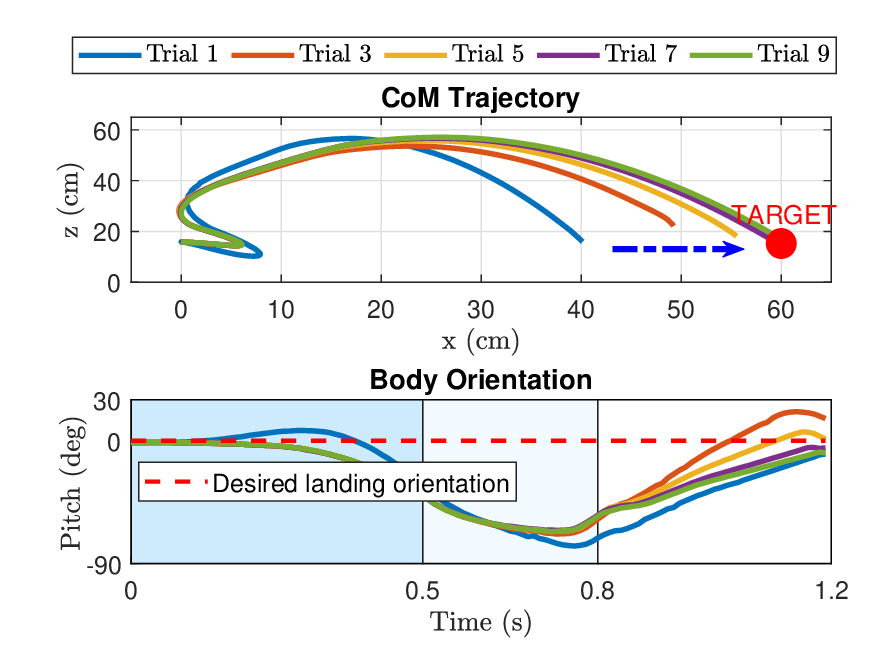}}

	}
	
	\caption{\textbf{Hardware: Learning to accomplish more challenging tasks from a simpler one.} The left figures show the learning process of jumping on the box by leveraging the jumping skills obtained from a simple jump of $40~cm$ (see Fig. \ref{fig:jump_40d0h_traj}). The robot collides with the box in the first trial, then refines its learning force control over time until successfully jumping on the box at $(x,z)=(50, 20) cm$. It only needs 7 trials to accomplish this challenging task with our proposed framework. The right figures show the iterative learning progress to make the robot jump further to the $60cm$ goal. The robot leaps $60cm$ successfully after $9$ trials. The body position and orientation of these learning processes are recorded at the bottom.}
	\label{fig:jump_from_40cm_hardware}
\end{figure*}

To further illustrate the efficiency, we compare our approach with other MPC frameworks in terms of target accuracy for jumping with a long flight phase (i.e. $400ms$). We select the jumping on box task at $(x,z)=(50,20)cm$ for comparison (see Fig. \ref{fig:mpc_and_ilc}). 
A full-body trajectory optimization is solved to obtain a jumping reference, which then is tracked by MPC. \add{To enable real-time planning, MPC often relies on model simplification (e.g. single rigid body dynamics) and assumes a limited prediction horizon \cite{YanranDingTRO, GabrielICRA2021,Quad-SDK,park2017high}).
However, to achieve target accuracy, MPC typically requires accurate predictions for the entire long flight
phase. Therefore, relying on a simplified model will affect the prediction accuracy for the entire flight, posing a
challenge for control to achieve target accuracy. As a result, the robot can only jump a distance of $40 \text{cm}$ and falls shortly in front of the box.} 
\add{This can be explained by the utilization of a simplified model for prediction, which normally affects the accuracy of state prediction for the whole flight phase.}
\add{Our approach, on the other hand, allows the robot to improve its target jumping accuracy over trials and finally accurately leap $50 \text{cm}$ and successfully land safely on the platform.}

\subsection{Simulation: Practice Simple to more challenging tasks}
We then verify the effectiveness of the proposed framework that enables the robot to accomplish more challenging tasks at high accuracy within several trials, by leveraging skills learned from a simple task. We start with a successful forward jump of $60~cm$ on the flat ground, then perform some challenging jumping tasks as follows:




\textit{Task I - Jumping on a high box}: 
The objective is to train a robot to jump onto a high box with a height of $h=30~cm$ from a distance of $d=60~cm$. Initially, the robot failed to jump on the box in its first attempt as it only managed to jump forward by $60~cm$ on the ground, as shown in Fig. \ref{fig:jump_simulation}a. However, with the help of iterative learning, we were able to enhance the robot's target tracking performance over time. As a result, the robot was able to jump on the box successfully after only three attempts, as illustrated in Fig. \ref{fig:jump_simulation}a. This demonstrates the effectiveness of our approach in generalizing the jumping task to a higher goal within a few trials.

\textit{Task II - Carrying unknown mass}: We aims to enable the robot jump forward $60cm$ while carrying a load of $2kg$, as shown in Fig. \ref{fig:jump_simulation}b. The load's weight, however, is unknown to the controller and is up to $20\%$ of the weight of the robot's trunk.
In the first trial, we tried to use the "memory torque and joint profile" from a prior successful jump without any load. However, the robot failed to jump the desired distance and could only jump a short distance of $32cm$ due to the unknown heavy load it was carrying.

By using the proposed ILC, we enable the robot to jump to the target $60cm$ accurately by only executing further $9$ jumps, as detailed in Fig.\ref{fig:jump_simulation}b. 
Our controller can compensate for model uncertainties caused by the unknown load mass, and enable a highly accurate target jump (distance error $\approx 0.5~cm$) with only a small number of trials.

We set the weight matrices as $\bm{Q}^u=10^{-5}diag\lbrace 1,1,1,1\rbrace$, and $\bm{Q}^e=diag \lbrace 1, 3, 3, 0.01, 0.01, 0.01 \rbrace$ in our ILC design. For the joint PD controller, we use $K_{p,joint}= 100$ and $K_{d,joint}=2$.
These weights and controller parameters are applied for all jumping tasks in simulation and hardware experiments.

\subsection{Hardware Experiments}


For the hardware experiments, we utilize the Optitrack motion capture system (MoCap) to estimate the position and orientation of the robot trunk, instead of relying on the robot's internal state estimation, which normally provides inaccurate estimations for the flight phase. The MoCap system comprises four Optitrack Prime $13W$ cameras, for accurate tracking with an update frequency of $1kHz$. The actuator limits and on-board battery parameters of the robot are listed in Table \ref{tab:motor_batterry}. The ILC is formulated as QP and is typically solved within a second for each trial.


We verify the effectiveness of our approach to enable robots to jump to various challenging targets in hardware from a successful simple task. To this end, we first implement the iterative learning to complete a baseline task of jumping forward $40cm$ (\textit{Task I}), for example. With muscle memory learned from this short jump, the robot then performs jumping to a more distant location (\textit{Task II}) and jumps on different high boxes (\textit{Task III}).


\textit{Task I - Baseline jumping task}: For this task, we aim to enable the robot to accurately jump forward $40cm$ on the flat ground. This task can be completed in two ways:
\begin{itemize}
    \item \textit{Hardware}: Directly learning to jump $40cm$ from scratch in hardware, leveraging all three learning stages.
    \item \textit{Simulation-to-Hardware}: Learning to jump to $40~cm$ from scratch in simulation with all three learning stages. With the "muscle memory" learned from simulation, the robot adopts the simple-to-complex procedure (see. Fig. \ref{fig:simple_to_complex_framework}) to gradually improve its jumping skills until the robot leaps $40~cm$ successfully in hardware.
\end{itemize}

Learning from scratch in simulation and hardware is described in Fig. \ref{fig:overview_framework}. The Stage I consists of 5 trials, followed by 5 trials of the Stage II. Finally, the robot performs the Stage III step until it reaches the goal. As we observe, learning from scratch directly in hardware takes $15$ trials to complete the task. 
For the second learning option (i.e., \textit{Simulation-to-Hardware}), when we transfer the muscle memory learned from simulation to hardware, the robot can jump only $19cm$ due to sim-to-real gap transfer. However, the robot only needs more $9$ trials to reach the final target $40cm$ in hardware, as shown in Fig. \ref{fig:jump_40d0h_traj}. We also observe a small landing angle of less than $10^0$ degrees for all jumps and just only $1^0$ at the last trial. Therefore, it proves the effectiveness of our approach to enable the use of some prior knowledge from learning in simulation to reduce the number of trials in hardware.

Once the jumping $40cm$ completed, we now discuss the capability of robot to conquer more challenging tasks as follows.

\NEW{
\begin{table}
  \begin{minipage}{1\linewidth}
    \centering
    \begin{tabular}{ *{3}{c} }
      \toprule
      \makecell{Jumping Tasks} & \makecell{\textbf{Learning from}\\
      \textbf{Scratch}\\
      Number of trials} & \makecell{\textbf{Learning by} \\
      \textbf{Our Method}\\
      Number of trials}\\
      \midrule
    \midrule
    Forward $50cm$ [Exp] & 15 & 3\\
    Forward $60cm$ [Exp] & 19 & 7\\
    Box jumping $(50,20)cm$ [Exp] & 19 & 9\\
    Box jumping $(60,10)cm$ [Exp] & 18 & 8 \\
    \midrule
    Box jumping $(50,40)cm$ [Sim] & 20 & 5 \\
    Box jumping $(60,30)cm$ [Sim] & 18 & 4 \\
    Carrying mass $2kg$ [Sim] & 20 & 9 \\
      \bottomrule
    \end{tabular}
    \caption{\textbf{Comparison between learning from scratch and our approach} - With our approach, the robot hardware applied what it learned from jumping forward $40cm$ to accomplish four more challenging targets. For simulation, we consider the forward jump of $60cm$ as a simple practice, then learn the other three challenging tasks. It validates that our method would significantly reduce the number of trials by around two to five times, compared with learning from scratch to finish all the tasks.}\label{tab:comparison_table}
  \end{minipage}%
\end{table}
}

\textit{Task II - Jumping to greater distances: } We firstly demonstrate the capability of robot jumping to a distant location at $60cm$ within a small number of trials.
Starting from the successful forward jump $40cm$ obtained in the \textit{Task I}, robot learns to reach the new target after $9$ jumps, while keeping a small landing angle during the whole learning progress, as shown in Fig.\ref{fig:jump_from_40cm_hardware}.

 \begin{figure}[!t]
    	\centering
    	{\centering
    		\resizebox{1\linewidth}{!}{\includegraphics[trim={0.5cm 0.7cm 0cm 0.1cm},clip]{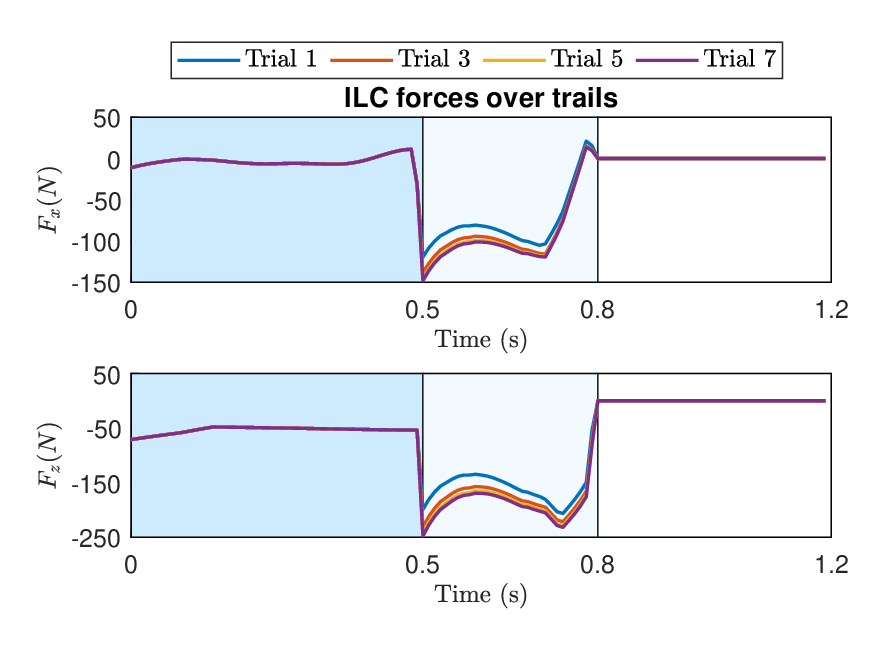}}}\\
    	\caption{\textbf{The contact force from ILC for jumping on box $\bm{(x,z)=(50,20)cm}$} (a) horizontal force for rear leg (b) vertical force for rear leg.} 
    	\label{fig:jump_onbox_GRF}
\end{figure} 

\begin{figure}[!t]
    \centering
    {\centering
        \resizebox{1\linewidth}{!}{\includegraphics[trim={0cm 1.4cm 0cm 0.2cm},clip]{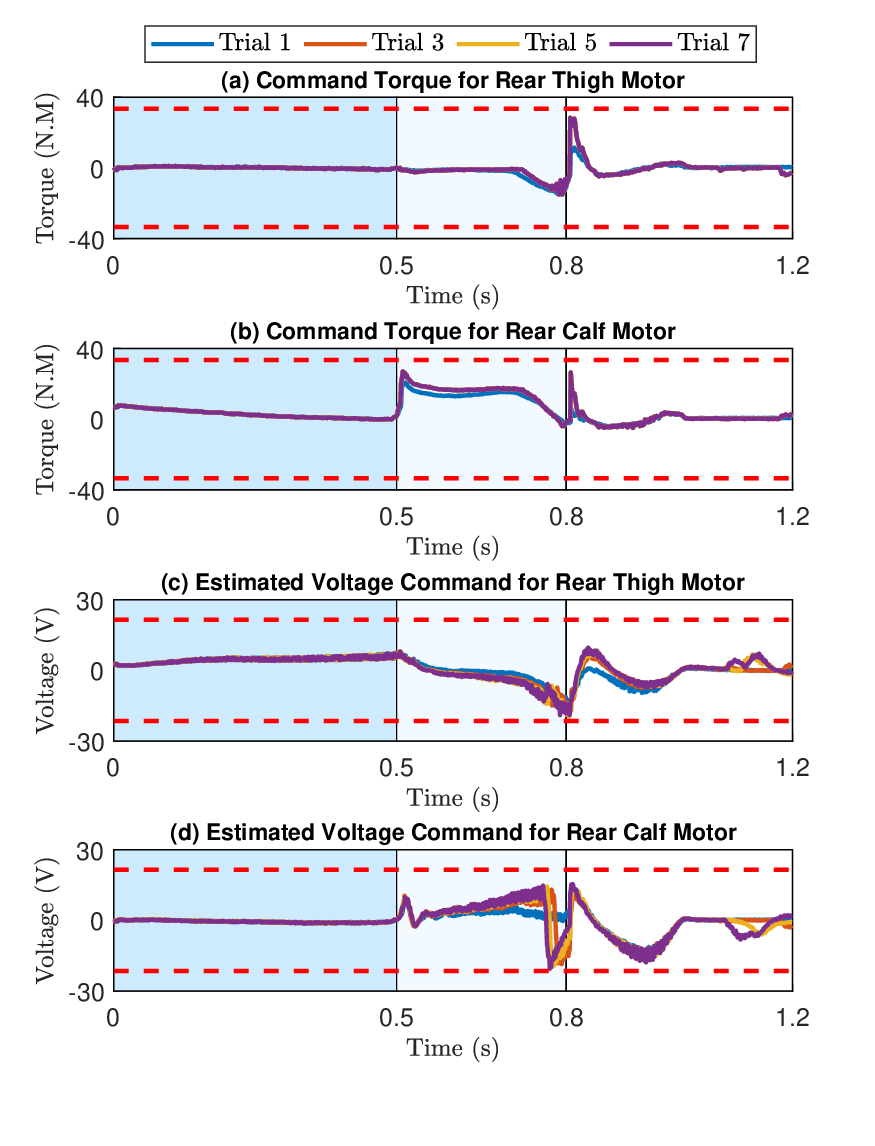}}}\\
    \caption{\textbf{Command torque and voltage over trials for jumping on box $\bm{(x,z)=(50,20)cm}$}. We examine the motors of the rear legs as jumping forward mainly utilizes power on these motors.  (a) Torque for the rear leg calf motor, (b) Torque for the rear leg knee motor, (c) Voltage for the rear leg calf motor, (d) voltage for the rear leg knee motor. The torque and the voltage all satisfy the limits $[-33.5,33.5](N.M)$ and $[-21.5,21.5](V)$, respectively.}
    \label{fig:jump_onbox_voltage}
\end{figure} 

\textit{Task III - Jumping on high boxes: } We now verify the method for accurate jumping on boxes at two different locations and height: $(x,z)=(50,20)cm$ and $(x,z)=(60,10)cm$, while respecting its hardware limits, i.e., torque limits and motor dynamic constraints.
The learning results for these tasks are verified in Fig.\ref{fig:jump_from_40cm_hardware} and Fig.\ref{fig:ILC_Introduction}b. We only consider the jumping to $(x,z)=(50,20)cm$ to further discuss here due to space limitation. In all iterative jumps to this target, we enforce friction cone limits, torque limits, and motor dynamics constraints to prevent slippery and enable successful hardware transfer. As shown in Fig. \ref{fig:jump_onbox_GRF}, the friction cone constraints are guaranteed with a friction coefficient of $\mu=0.6$. The Fig. \ref{fig:jump_onbox_voltage} shows that the total command torques $\bm{\tau}_{total}$ satisfies the torque motor limits $\bm{\tau}_{max}=33.5(Nm)$, and the estimated command voltages are within the battery's voltage supply $V_{max}=21(V)$.

The Table. \ref{tab:comparison_table} summarizes the hardware experiments and simulation for various targets. We take a comparison between jumping from scratch and our approach. \add{With learning from scratch, the process consists of three sequential optimization stages. On the other hand, our approach allows the robot to optimize only the last stage (Stage III) when a more challenging task is given}. With our approach, the robot can reduce necessary trials by around two to five times to complete all the same jumping tasks.
   


\section{Conclusion}
\label{sec:ILC_Conclusion}


In this paper, we have presented an ILC approach that enables the robot to master its jumping skills for challenging tasks through simple practices.
Through extensive experiments, we have verified that the proposed method enables the robot to gradually refine its jumping technique over trials. The robot finally reaches desired targets accurately and safely, despite the presence of model uncertainties. Importantly, the robot leverages its prior successful experience to accelerate learning progress in accomplishing many challenging tasks, without the need to learn from scratch.



Our approach relies on the use of single rigid body dynamics, which is widely used for modeling legged robot systems. Therefore, the method can be generic enough to be extended to other types of legged robots with any number of legs.
Our proposed framework lays the foundation for our future work, which includes expanding the scope of our method to 3D jumping motions and exploring other types of robots, such as humanoids and wheel-legged robots.

\add{
Our current framework assumes the same initial conditions for all learning trials. In the future, we extend this to scenarios involving uncertainties and disturbances introduced at the initial conditions. One promising approach to mitigating such challenges is adaptive iterative learning control (ILC). This method can enhance the performance and robustness of control systems by dynamically adjusting parameters based on feedback from previous iterations and real-time observations (e.g.\cite{Miao_adaptive_ILC_nonlinearsystem}). Another limitation we observe is that the trajectory can converge slowly when the robot reaches close to the goal points. A potential solution in our future work is to leverage the learning progress from more historical trials, rather than relying solely on the immediate last trial, to achieve faster convergence.}

\section{Appendix}\label{sec:appendix}

\begin{table}[bt!]
	\centering
	\caption{System and on-board battery parameters \cite{A1}}
	\begin{tabular}{cccc}
		\hline
		Parameter & Value & Units\\
		\hline
  		Body and Link 
    Length &   0.366, 0.2   &  m \\[.5ex]
            Body, Thigh, Calf Weight &   9.60, 1.61, 0.66   &  kg \\[.5ex]
		Max Joint Torque  &  33.5 & ${Nm}$  \\[.5ex]
		Max Joint Speed    & 21  & ${Rad}/{s}$  \\[.5ex]
            Max Battery Voltage &   21.5   & ${V}$   \\[.5ex]
		\hline 
		\label{tab:motor_batterry}
	\end{tabular}
  \vspace{-2.2em}
\end{table}

\subsection{Simplified dynamics for quadruped jumping} \label{subsec:dynamics}

We revisit a simplified form of rigid body dynamic in the vertical plane as follows \cite{park2017high, YanranDingTRO}:
    \begin{align} \label{eq:ILC_simplified dynamics}
    \ddot{\bm{p}} =\frac{\sum_{i=1}^{n_l}\bm{u}_{i}}{m}-\bm{g}, ~~~ \frac{\bm{d}}{d t}(I {\bm{\omega}}) =\sum_{i=1}^{n_l} \bm{r}_{i} \times \bm{u}_{i},
    \end{align}

where $\bm{p}$ is the CoM position in the world frame; $\bm{r}_i=[r_{ix};r_{iz}]$ and $\bm{u}_i=[u_{ix};u_{iz}]$ denotes the position of contact point relatively to CoM, and contact force of foot $i^{th}$ respectively in the world frame; $\bm{\omega}=\dot{\theta} \vec{k}$ is angular velocity of the body; $\theta$ is a pitch angle. 
We consider the system states as $\bm{x}=[\bm{p};\theta;\dot{\bm{p}};\dot{\theta}] \in \mathbb{R}^6$.

In iterative learning formulation, we denote $k$ as the index of trial, and $t$ represents for time step in each trial. Then, the discrete time representation of jumping dynamics for trial $k$ can be written as (first order Taylor estimation):
\begin{subequations}
    \begin{align}
        \bm{x}_{k,t+1} &=\underbrace{\left(\bm{I} + \delta_t\bm{A}_c\right)}_{\bm{A}}  \bm{x}_{k,t}+ \underbrace{\delta_t \bm{B}_c (\bm{r}_1, \bm{r}_2)}_{\bm{B}_{k,t}} \bm{u}_{k,t}, \label{eq:simplified robot dynamics_discrete_ilc_1}\\
        \bm{y}_{k,t} & =\bm{x}_{k,t} \label{eq:simplified robot dynamics_discrete_ilc_2}
    \end{align}
\end{subequations}
where
\begin{subequations}
    \begin{align}
            & \bm{A}_c = \begin{bmatrix}
                    \bm{0} & \bm{I} & \bm{0}\\
                    \bm{0} & \bm{0} & \bm{\epsilon}_g\\
                    \end{bmatrix}, 
            \bm{B}_c = \begin{bmatrix}
                    \bm{0} & \bm{0}\\
                    \frac{\bm{I}}{m} & \frac{\bm{I}}{m} \\
                    \sigma_1 & \sigma_2 \\ 
                        \end{bmatrix}, \\
            & \bm{\epsilon}_g= \begin{bmatrix} 0 & -1 & 0
\end{bmatrix}^\top, \sigma_i= I^{-1} \begin{bmatrix} -r_{iz} & r_{ix}
\end{bmatrix}
    \end{align}
\end{subequations}
\subsection{Low-level controller} Our designed ILC is the force-based controller. The contact force directly affects the body trajectory as described by SRB dynamics. 
During trial $k+1$, the optimal contact force $\bm{u}_{k+1}$ will be converted to the optimal torque $\bm{\tau}_{ILC}$ via
\begin{equation}
    \bm{\tau}_{ILC} = \bm{J}(\bm{q}_{k+1})^\top \bm{R}_{k+1}^\top \bm{u}_{k+1},
\end{equation}
The ILC torque will then be combined with the joint feedback controller $\bm{\tau}_{PD}$ to generate a total command torque $\bm{\tau}_{total}$ applying to the robot motors:
\begin{subequations}
\begin{align}
    \bm{\tau}_{total} &= \bm{\tau}_{PD} + \bm{\tau}_{ILC},\\
    \bm{\tau}_{PD} &= \bm{K}_{p,joint} (\bm{q}_d -\bm{q}) + \bm{K}_{d,joint}  (\bm{\dot{q}}_d -\bm{\dot{q}}),
    \label{eqn:jumping_full}
\end{align}
\end{subequations}
where $\bm{K}_{p,joint}$ and $\bm{K}_{d,joint}$ are diagonal matrices of proportional and derivative gains in the joint coordinates. The joint PD controller $\bm{\tau}_{PD}$ running at 1 \emph{kHz}. Reference joint angles $(\bm{q}_d)$, joint velocities $(\bm{\dot{q}}_d)$ are obtained from trajectory optimization with sampling time of 10 \emph{ms}. They are then linearly interpolated to 1 \emph{ms}. 
It is noted that in the first trial, we utilize the ground contact force reference $\bm{u}_d \triangleq \bm{f}_d$ obtained from trajectory optimization (as detailed in Section \ref{sec: ILC_TO}). In particular, we execute the following torque command for the first trial:
\begin{equation}
    \bm{\tau}_{total} = \bm{\tau}_{PD} + \bm{J}(\bm{q})^\top \bm{R}^\top \bm{u}_d,
\end{equation}

